\documentclass[10pt, a4paper]{article}

\usepackage[]{lrec-coling2024} 

\usepackage{amsmath,amsfonts}
\usepackage{tcolorbox}
\usepackage{CJK}
\usepackage{bm}
\usepackage{array}
\usepackage{subfig}
\usepackage{multirow}
\usepackage{arydshln}
\usepackage{graphicx}
\usepackage{booktabs}
\usepackage{caption}

\title{Pluggable Neural Machine Translation Models\\via Memory-augmented Adapters}

\name{Yuzhuang Xu$^{1,\dagger}$\thanks{$^{\dagger}$Equal contribution}, Shuo Wang$^{1,\dagger}$, Peng Li$^{2,\ast}$, Xuebo Liu$^{3}$\\
{\bf \large Xiaolong Wang$^{1}$, Weidong Liu$^{1,4}$, Yang Liu$^{1,2,\ast}$\thanks{$^\ast$Corresponding authors}}}

\address{$^{1}$Department of Computer Science \& Technology, Tsinghua University, Beijing, China \\
$^{2}$Institute for AI Industry Research (AIR), Tsinghua University, Beijing, China \\
$^{3}$Harbin Institute of Technology, Shenzhen, China\\
$^{4}$Zhongguancun Laboratory, Beijing, China \\
\texttt{\{xyz21thu,wangshuo.thu\}@gmail.com, lipeng@air.tsinghua.edu.cn}\\
\texttt{liuyang2011@tsinghua.edu.cn}\\
}

\abstract{
Although neural machine translation (NMT) models perform well in the general domain, it remains rather challenging to control their generation behavior to satisfy the requirement of different users. Given the expensive training cost and the data scarcity challenge of learning a new model from scratch for each user requirement, we propose a memory-augmented adapter to steer pretrained NMT models in a pluggable manner. Specifically, we construct a multi-granular memory based on the user-provided text samples and propose a new adapter architecture to combine the model representations and the retrieved results. We also propose a training strategy using memory dropout to reduce spurious dependencies between the NMT model and the memory. We validate our approach on both style- and domain-specific experiments and the results indicate that our method can outperform several representative pluggable baselines. Code and data are available at \url{https://github.com/xuyuzhuang11/StyleMT}
 \\ \newline  }

\begin{document}

\maketitleabstract

\section{Introduction}
\label{sec:introduction}

In recent years, modern neural machine translation (NMT;~\citealp{Vaswani:2017:Transformer}) systems are often developed with large-scale parallel data extracted from the Web~\cite{Liu:2020:mBART,Fan:2021:m2m100}, whose style and content are driven by the average distribution of data from many domains~\cite{Vu:2021:Customize}. Therefore, the performance of strong NMT models is close to or even better than human translators in the general domain~\cite{Hassan:2018:Human,Kocmi:2022:WMT}.

However, MT customers may have some special requirements, including both style- and domain-specific individual demands~\cite{Michel:2018:PersonNMT,Zhang:2022:PersonNMT}. For instance, some users may want translations in a special style, while some others may need to translate medical texts. These requirements can be quite diverse among different customers and retraining or fine-tuning the model for each user entails significant development costs, especially with limited data from users.

\begin{figure}[t]
  \centering
  \includegraphics[width=0.48\textwidth]{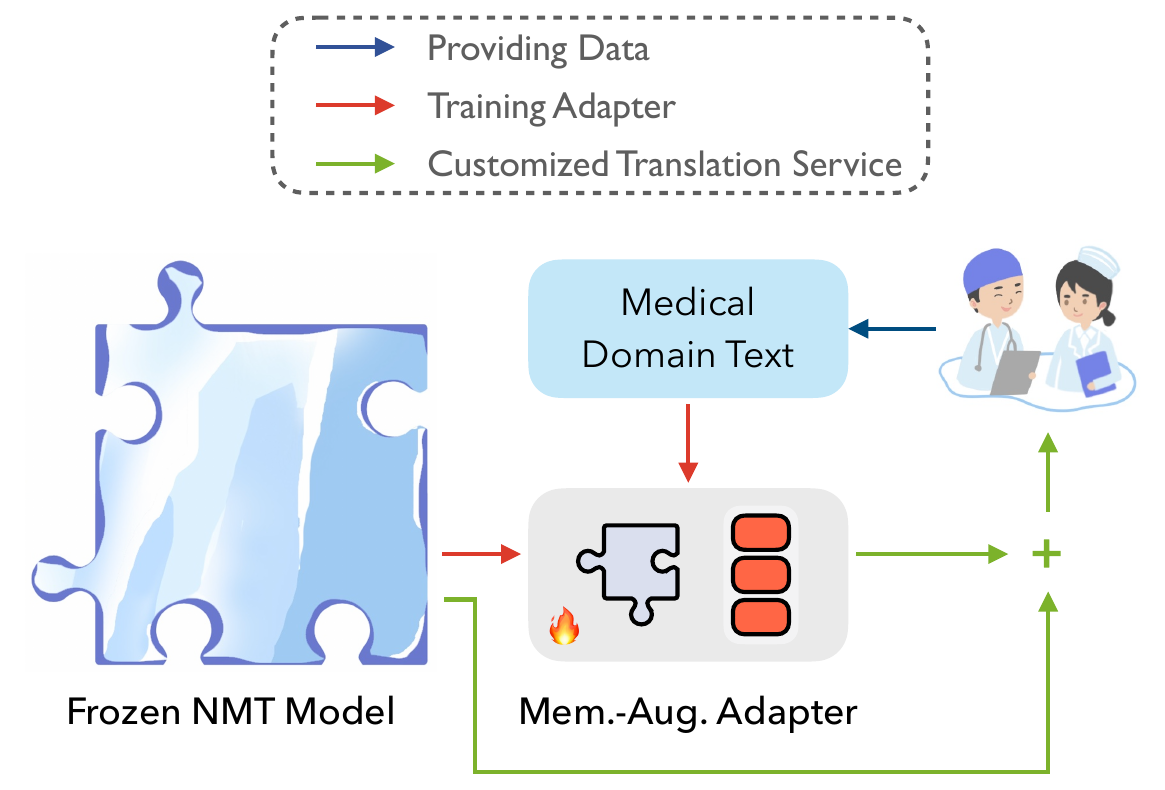}
  \caption{
    A frozen and pluggable NMT model using memory-augmented plugins. For each user group with special requirements, we can develop a plugin for them without affecting other users.
  }
\label{fig:intro}
\vspace{-1.0em}
\end{figure}

Fortunately, {\em pluggable} methods~\cite{Keskar:2019:CTRL,Dathathri:2020:PPLM,He:2021:EfficientKNN} bring hope to handle the aforementioned user requirements, which employ additional modules to steer pretrained models.
As shown in Figure~\ref{fig:intro}, the users can provide some text samples for the NMT model to imitate.
We will then learn a plugin to control the NMT model to satisfy the user demands without optimizing the parameters in the original model, by which we can maintain the performance of the pretrained model, alleviating the risk of catastrophic forgetting~\cite{Kirkpatrick:2017:CF}.

Some researchers suggest lightweight parametric plugins for controlling pretrained models~\cite{Houlsby:2019:Adapter,Bapna:2019:Adapter,Pfeiffer:2021:Adapterfusion,Ruckle:2021:Adapterdrop,Li:2021:Prefix,Mao:2022:Unipelt}. For machine translation, such plugins can also tailor model behavior for diverse user demands.
However, recent studies find that there exists a performance bottleneck of fully-parametric pluggable methods~\cite{Bapna:2019:Adapter,Li:2021:Prefix,Ding:2022:DeltaTuning}: increasing the number of trainable parameters can not always lead to better performance.
Inspired by the recent progress of retrieval-augmented models~\cite{Lewis:2020:Retrieve,Khandelwal:2020:kNNLM,Khandelwal:2021:kNNMT,He:2021:EfficientKNN,He:2021:Fast}, we propose to increase the expressive power~\cite{Li:2021:Prefix} of parametric plugins through external memories, which we term it {\em memory-augmented adapter}.

The main challenges of the memory-augmented adapter are two-fold: (1) {\em constructing} memories that can provide useful customization information; and (2) {\em integrating} the memories into existing NMT models without quality loss. Although long phrases can provide more contextualized information, matching long sequences between queries and memory items is more difficult than matching shorter ones. We propose to build multi-granular memories to balance the amount of contextualized information and the retrieval difficulty.
Unlike many previous works~\cite{Khandelwal:2020:kNNLM,Khandelwal:2021:kNNMT,He:2021:EfficientKNN} that encode the source sentence and the target prefix as the key and the next token as the value, our memory can provide multi-scale translation knowledge~\cite{Li:2022:multiscale} that is suitable for queries coming from different layers of the NMT model~\cite{Hewitt:2019:Probe}. For memory integration, we propose a \textit{new adapter architecture} to better interpolate the original model representation and the retrieved vectors. Moreover, we propose a new training strategy with memory dropout to reduce spurious dependencies between the NMT model and the provided memory.
We conduct experiments for both style- and domain-related customizations and the results show the superiority of our method over many representative baselines.

\section{Related Work}
\label{sec:related}

\subsection{Style / Domain Adaptation for NMT}

Adapting NMT models to specific style or domain texts has been investigated in several previous works~\cite{Luong:2015:Domain,Niu:2017:Style,Chu:2018:Domain}.
For stylized NMT, many previous works focus on the formality control of translations~\cite{Niu:2018:Style,Wu:2021:Style}, of which the style has a clear definition. Most existing works need to train a specific model for each style. For instance,  \citet{Niu:2020:Style} mix the training data of both style transfer and machine translation to learn a formality-sensitive NMT model.
Given that the user-provided styles can be of great diversity, we aim to satisfy different style demands in a pluggable manner.

For domain adaptation,  \citet{Luong:2015:Domain} propose an effective method that fine-tunes an out-of-domain model with small-sized in-domain supervised corpora. \citet{Hu:2019:Domain} further design an unsupervised method, since parallel data is hardly available in many domains. \citet{Zheng:2021:Non-parametric} extend $k$NN-MT~\cite{Khandelwal:2021:kNNMT} to perform unsupervised domain adaptation. Our work is different from \citet{Zheng:2021:Non-parametric} in both memory design and usage and the experiments show that our proposed framework performs better than their approach.

\subsection{Machine Translation Customization}

Machine translation customization aims to satisfy the special requirements of different users. \citet{Vu:2021:Customize} propose to select data that is similar to the user-provided text samples and then train or fine-tune an NMT model for the corresponding user. Following \citet{Michel:2018:PersonNMT}, we believe that MT customization has some specific traits that distinguish it from common style and domain adaptation settings:
(1) The number of customization requirements is very large due to the personal variation among different MT system users;
(2) The available data is often very limited (even monolingual, let alone parallel) for each customization requirement.
Thus, we propose to leverage pluggable methods to customize existing NMT models.

\subsection{Pluggable Pretrained Models}


Pluggable methods aim to control the generation behavior of pretrained models without optimizing model parameters~\cite{Dathathri:2020:PPLM,Yang:2021:FUDGE,Liu:2021:Dexperts}. Some works propose to use parametric plugins, with \citet{Bapna:2019:Adapter} and \citet{Houlsby:2019:Adapter} inserting some adapters, \citet{Li:2021:Prefix} prepending some trainable vectors, and \citet{Hu:2022:LoRA} leveraging low-rank decomposition of matrices. Retrieval-augmented models, also treated as pluggable methods, augment the model with non-parametric memory. $k$NN-MT~\cite{Khandelwal:2021:kNNMT} combines the model prediction and retrieval distribution at the output layer. \citet{Chaudhuri:2022:Retro} build a chunk-level memory for language modeling. \citet{Chen:2022:Augmenting} encode questions and answers into key-value pairs for question answering. In this work, we aim to combine the merits of both parametric and non-parametric plugins and propose a new type of memory for NMT, which explicitly considers the phrases of different granularities.





\section{Background}

\subsection{Transformer Model}

We first give a description of some components in the Transformer~\cite{Vaswani:2017:Transformer} model. 
Given the input sentence $\mathbf{x}$, Transformer maps it into vectors via an encoder:
\begin{equation}
    \mathbf{E} = \mathrm{encoder}(\mathbf{x})
    \label{eq:encoder}
\end{equation}

where $\mathbf{E}\in\mathbb{R}^{|\mathbf{x}| \times d}$ and $d$ is the hidden size of the model. The encoder output is then utilized by the decoder, which is a stack of several independent layers. We use $\mathbf{D}^{(i)}$ to denote the output of the $i$-th decoder layer. Specifically, each decoder layer firstly employs a self-attention module to model the dependency between the target-side words:
\begin{equation}
\begin{split}
    \mathbf{S}^{(i)} &= \mathrm{attn} (\,
        \mathbf{D}^{(i-1)}, \mathbf{D}^{(i-1)}, \mathbf{D}^{(i-1)}
    \,) \\
    \mathbf{L}_1^{(i)} &= \mathrm{layernorm} (\,
        \mathbf{D}^{(i-1)} + \mathbf{S}^{(i)}
    \,)
\end{split}
\label{eq:decoder-selfattn}
\end{equation}
where $\mathrm{attn}(\mathbf{Q}, \mathbf{K}, \mathbf{V})$ is the multi-head attention and $\mathrm{layernorm}$ is the layer normalization.

After that, a cross-attention module is adopted to integrate the source-side information:
\begin{equation}
\begin{split}
    \mathbf{C}^{(i)} &= \mathrm{attn} (\,
        \mathbf{L}_1^{(i)}, \mathbf{E}, \mathbf{E}
    \,) \\
    \mathbf{L}_2^{(i)} &= \mathrm{layernorm} (\,
        \mathbf{L}_1^{(i)} + \mathbf{C}^{(i)}
    \,)
\end{split}
\label{eq:decoder-crossattn}
\end{equation}

The output of the cross-attention module is then projected with a feed-forward layer. The decoder output is finally used to estimate the probability $P(\mathbf{y} | \mathbf{x}; \bm{\theta})$, where $\mathbf{y}$ is the target sentence and $\bm{\theta}$ denotes the set of model parameters.

\subsection{Style Customization in NMT}

Similar to generating images with specific styles \cite{jing2022learning, ruiz2023dreambooth}, style customization in NMT means outputting translations with user-specified styles \cite{Michel:2018:PersonNMT, Syed:2020:Adapting}. For example, we want to generate translations in Shakespeare style in a Zh-En translation task. A simple example is as follows.

\begin{tcolorbox}
\begin{tabular}{@{}l@{\hspace{3pt}}p{0.8\linewidth}}
\textbf{Zh:} & \begin{CJK}{UTF8}{gbsn}哦上帝啊，请赐予我力量吧！\end{CJK} \\
\textbf{En (G):} & Oh God, please grant me strength! \\
\textbf{En (S):} & Oh \underline{Lord}, do \underline{thou} endow me with \underline{thy} might! \\
\end{tabular}
\end{tcolorbox}

\noindent ``\textbf{En (G)}'' denotes the output of vanilla translation model, and ``\textbf{En (S)}'' denotes the output of style-customized translation model using Shakespeare corpus. The underlined expressions are typical representations of Shakespeare style.

The task most closely related to style customization in NMT is author-stylized rewriting~\cite{Syed:2020:Adapting, singh2021drag}, which aims to rewrite a given text in the style of a specific author. ~\citet{Syed:2020:Adapting} sum up the user or author style into three levels, namely surface level, lexical level, and syntactic level styles~\cite{Syed:2020:Adapting}. These levels capture subtle differences in punctuation, word usage, and even sentence construction unique to individual authors, thereby making author-stylized rewriting a challenging task. Style customization in NMT not only shares the same challenges as author-stylized rewriting, but it must also simultaneously translate the provided text into the target language, presenting its own unique challenges.

\section{Approach}

\begin{figure*}[ht]
    \centering
    \begin{minipage}{0.64\textwidth}
        \centering
        \subfloat[Parallel text segments at different levels of granularities.]{\includegraphics[width=\textwidth]{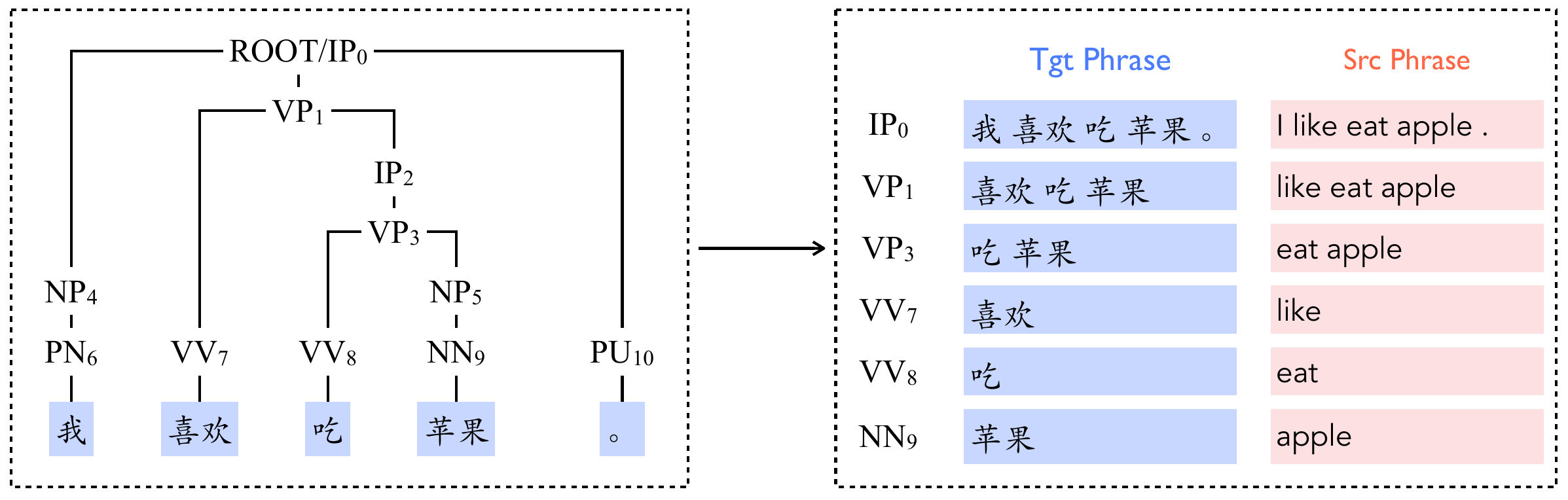}\label{fig:sub_build1}}
        
        \subfloat[Construction of multi-granular continuous memory.]{\includegraphics[width=\textwidth]{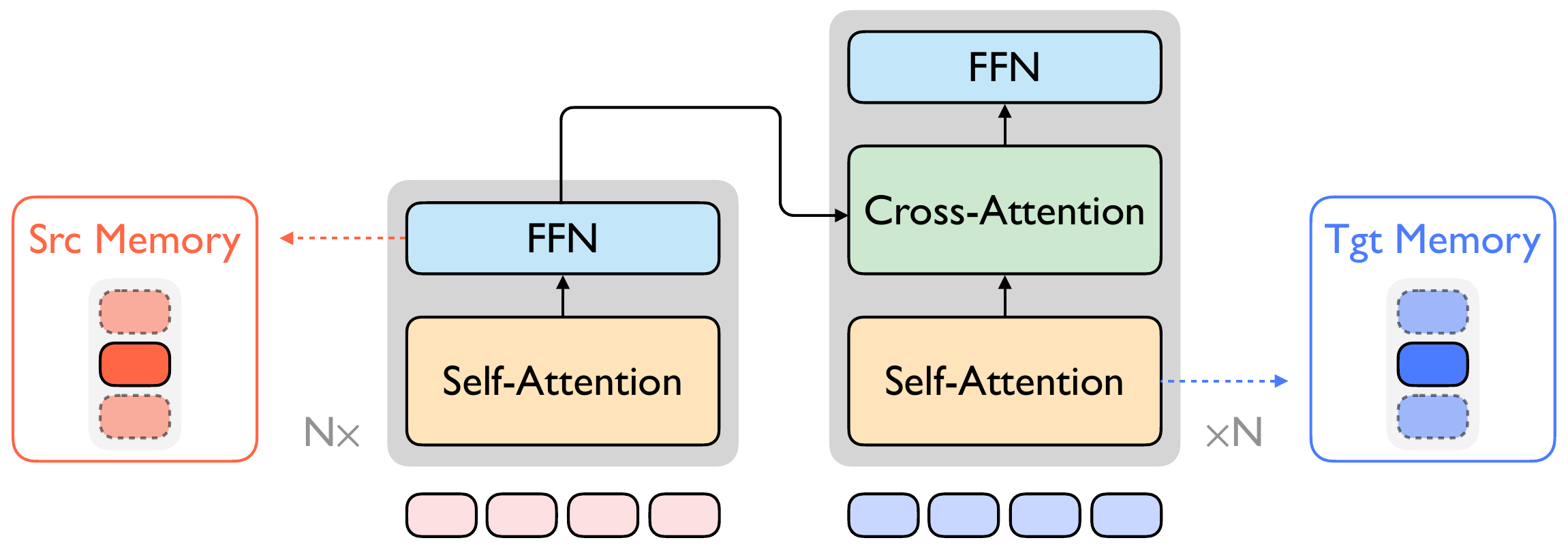}\label{fig:sub_build2}}
    \end{minipage}%
    \begin{minipage}{0.35\textwidth}
        \centering
        \subfloat[Adapter integration.]{\includegraphics[width=0.85\textwidth]{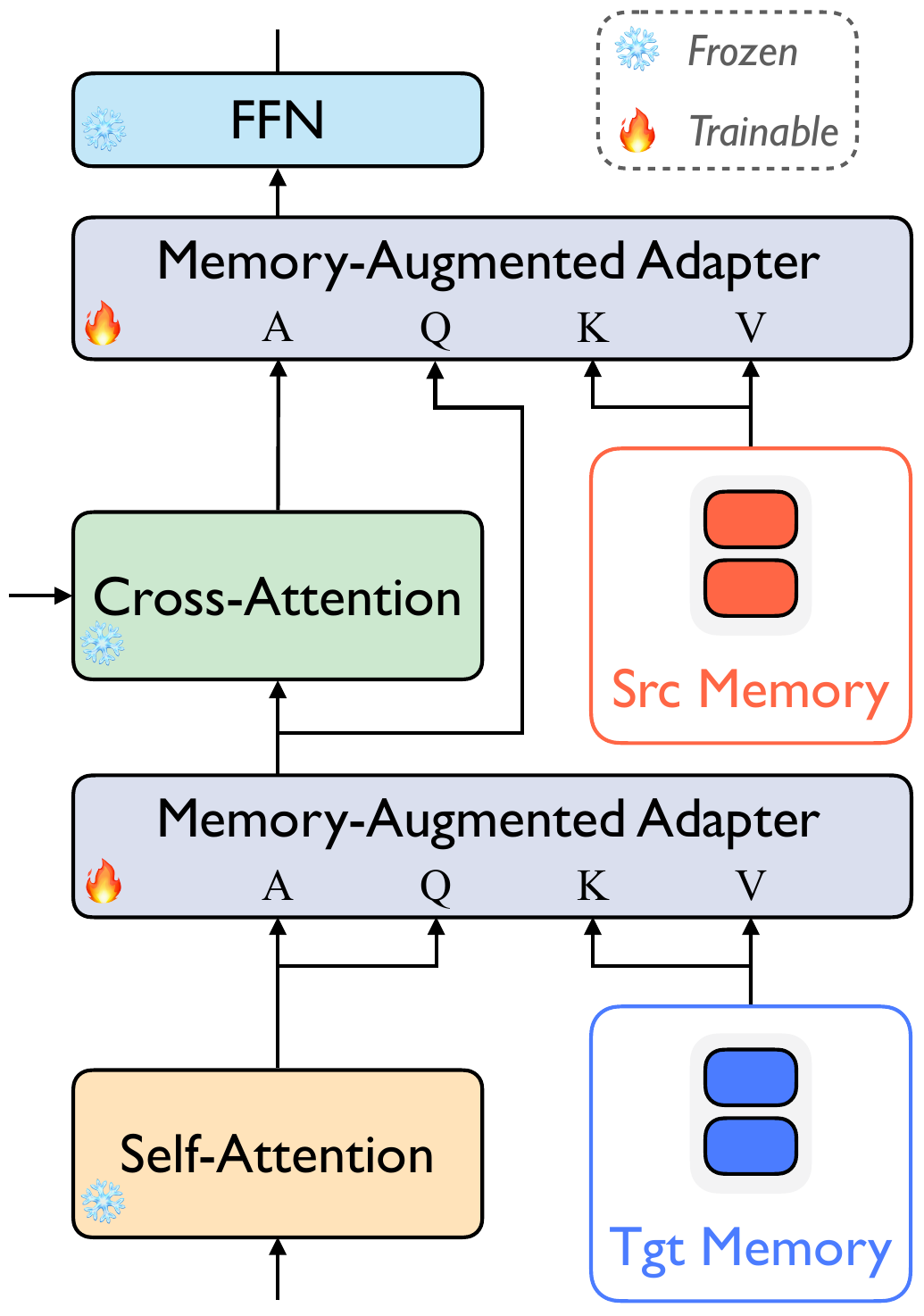}\label{fig:sub_memread}}
    \end{minipage}
    \caption{Construct and integrate memories. (a) We leverage parse trees to obtain multi-granular phrases. Each monolingual phrase is then translated by NMT models. (b) For each phrase pair, we perform a forward computation in the teacher-forcing manner and record some intermediate representations into the memory. (c) Illustration of adapter integration. The adapter retrieve and leverage the memories.
    }
    \label{fig:memory-main}
\end{figure*}



\subsection{Overview}

In this work, we aim to enable NMT users to control existing NMT models by simply providing some examples. To this end, we propose a memory-augmented adapter to help NMT models imitate the user-provided text samples.
Specifically, we propose the multi-granular memory that can better leverage multi-scale patterns, which have proven to be important for NMT~\cite{Li:2022:multiscale}. We also propose a new adapter (i.e., memory-augmented adapter) to integrate external memory into NMT models. We will explain how to construct and utilize the memory in the following two subsections.

\subsection{Multi-granular Continuous Memory}
\label{sec:memory-construction}

Our memory needs not only to extract essential information from user-provided text but also to be easy to retrieve for models. For the first objective, we propose to build the memory with parallel phrase pairs, which reflect the translation pattern required by the customer. However, it is non-trivial to determine the granularity of the used phrases. Storing only short ones may waste a lot of contextualized information while storing too many long phrases would make it rocky to match the query and the memory items. To address this issue, we propose to construct a multi-granular memory to balance the amount of contextualized information and the retrieval difficulty. 
As shown in Figure~\ref{fig:sub_build1}, we use parse trees to extract multi-granular phrases, which can identify more meaningful boundaries than random splitting.
The extracted phrases are then translated by NMT models to form parallel phrase pairs.

For the second objective, we propose to use the same model to build and utilize the memory. For each phrase pair, we perform forward computation to get the continuous representation at each layer of the involved NMT model. We store the encoder output $\mathbf{E}$ as the source-side memory and the self-attention output $\mathbf{S}^{(i)}$ at every decoder layer as the target-side memory. See Eq.~(\ref{eq:encoder}) and (\ref{eq:decoder-selfattn}) for more details of the stored representations. Each memory item is averaged among the representations of all tokens in a phrase, whose size is $d$. Figure~\ref{fig:sub_build2} shows an example. The reason we extract $\mathbf{E}$ and $\mathbf{S}^{(i)}$ as our memory is that these representations are at the same layer where we perform memory retrieval. Our motivation is to narrow down the gap between the memory items and the queries,
making it easier for the model to read the memory.

We focus on using monolingual user-provided data in this work since parallel data is often unavailable for most requirements. However, our method can be easily extended for bilingual data, from which we can automatically extract phrase pairs based on unsupervised word alignment algorithms~\cite{Dyer:2013:fastalign,Chen:2021:maskalign}.

\subsection{Memory-augmented Adapter}
\label{sec:memory-augmented-adapter}


\begin{figure}[t]
    \centering
    \includegraphics[width=0.32\textwidth]{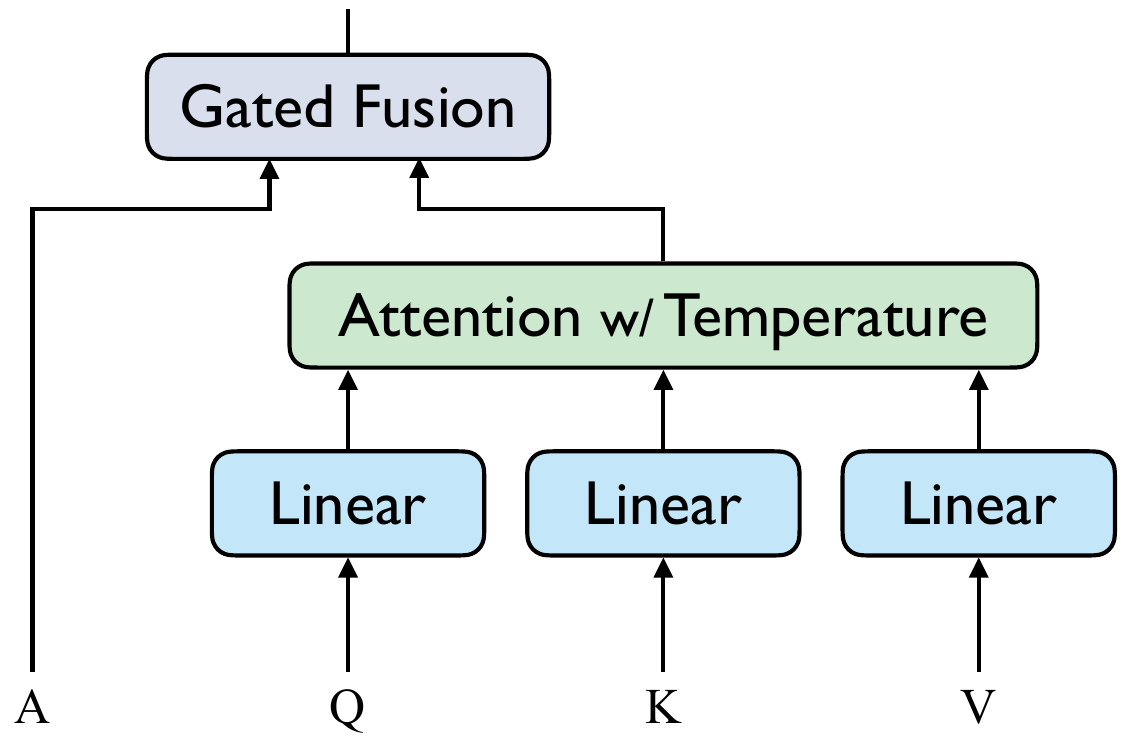}
    \caption{
        Memory-augmented adapter architecture.
    }
\label{fig:memory-aug-adapter}
\end{figure}

\paragraph{Adapter Architecture}
We propose a new type of adapter to read memory.
The memory-augmented adapter has 4 inputs: anchor, query, key, and value, which can be represented as $\mathbf{A}$, $\mathbf{Q}$, $\mathbf{K}$, and $\mathbf{V}$. Anchors and queries are derived from the frozen NMT model while keys and values come from the memory. As depicted in Figure~\ref{fig:memory-aug-adapter},
we use an attention module to generate the retrieved result:
\begin{equation}
\label{eq:retrieval-attention}
\begin{split}
    \mathbf{R} = \mathrm{softmax} (\,
    \mathbf{Q}\mathbf{W}_q\mathbf{W}_k^{\top}\mathbf{K}^{\top} /\, T
     \,)\mathbf{\mathbf{V}\mathbf{W}}_{v}
\end{split}
\end{equation}
where $\mathbf{W}_{q}, \mathbf{W}_{k}, \mathbf{W}_{v} \in \mathbb{R}^{d \times d}$ and the retrieved result $\mathbf{R}$ has the same shape with $\mathbf{Q}$. $T$ is a hyperparameter to control the sharpness of the retrieval distribution.
To avoid the model being completely dependent on the retrieved result $\mathbf{R}$ that can be erroneous in some cases, we also take in an anchor from the original model, which is combined with $\mathbf{R}$ via a gated fusion module:
\begin{equation}
\begin{split}
    \lambda &= \mathrm{sigmoid} (\,\mathrm{relu}(\,[\mathbf{A};\mathbf{R}]\mathbf{W}_1\,)\mathbf{W}_2\,) \\
    \mathbf{O} &= \lambda\, \mathbf{A} + (1 - \lambda)\, \mathbf{R}
\end{split}
\end{equation}
where $\mathbf{O}$ is the adapter output, which has the same shape as the anchor $\mathbf{A}$. $\mathbf{W}_1 \in \mathbb{R}^{2d \times d}$ and $\mathbf{W}_2 \in \mathbb{R}^{d \times 1}$. $\lambda$ is the learned interpolation ratio.

\paragraph{Adapter Integration} We apply the memory-augmented adapter to the self- and cross-attention modules in the decoder, since these two components are important for target-side language modeling and source-side information utilization. At the $i$-th decoder layer, we use the self-attention output $\mathbf{S}^{(i)}$ as queries to read the target-side memory:
\begin{equation}
\begin{split}
    \mathbf{O}^{(i)}_1 = \mathrm{memadapt}(\,
        \mathbf{S}^{(i)}, \mathbf{S}^{(i)}, \mathbf{M}_t^{(i)}, \mathbf{M}_t^{(i)}
    \,) \\
\end{split}
\end{equation}
where $\mathrm{memadapt}(\mathbf{A}, \mathbf{Q}, \mathbf{K}, \mathbf{V})$ denotes the memory-augmented adapter. $\mathbf{M}_t^{(i)} \in \mathbb{R}^{N_t^{(i)} \times d}$ represents the target-side memory, where $N_t^{(i)}$ denotes the number of items in $\mathbf{M}_t^{(i)}$. The adapter output $\mathbf{O}_1$ is then provided to the layer normalization module:
\begin{equation}
    \mathbf{L}_1^{(i)} = \mathrm{layernorm} (\,\mathbf{D}^{(i-1)} + \mathbf{O}_1^{(i)}\,)
\end{equation}

Similarly, we read the source-side memory in the cross-attention module:
\begin{equation}
\begin{split}
    \mathbf{O}_2^{(i)} &= \mathrm{memadapt}(\,
    \mathbf{C}^{(i)}, \mathbf{L}_1^{(i)}, \mathbf{M}_s^{(i)}, \mathbf{M}_s^{(i)}
    \,) \\
    \mathbf{L}_2^{(i)} &= \mathrm{layernorm}(\,\mathbf{L}_1^{(i)} + \mathbf{O}_2^{(i)}\,)
\end{split}
\end{equation}

Figure~\ref{fig:sub_memread} shows an example. To reduce the redundancy that a phrase pair would repeatedly appear in memories at every decoder layer, we split all the phrase pairs into $L$ parts, where $L$ is the number of decoder layers. Each layer only stores one part of phrase pairs.


\paragraph{Training Strategy} Inspired by dropout~\cite{Srivastava:2014:Dropout} that can effectively reduce spurious co-adaptation between model parameters, we propose a memory dropout approach to prevent NMT models from being too dependent on some specific memory items. When training the memory-augmented adapter, we randomly drop part of the memory items. Let $\mathbf{M}$ be the full memory and $\hat{\mathbf{M}}$ be the remained memory after memory dropout, the overall loss can be given by
\begin{equation}\label{eq:totloss}
    \raisebox{-0.9em}{
    \makebox[0.5\columnwidth]{$
        \begin{aligned}
            \mathcal{L} &=\quad \underbrace{\mathcal{L}_{\mathrm{NLL}} (\,P(\mathbf{y} |\mathbf{x}, \bm{\theta}, \mathbf{M})\,)}_{\text{loss using full memory}}\\ 
            &+\quad \underbrace{\alpha\,\mathcal{L}_{\mathrm{NLL}} (\,P(\mathbf{y} |\mathbf{x}, \bm{\theta}, \hat{\mathbf{M}})\,)}_{\text{loss using dropped memory}} \\
            &+\quad  \underbrace{\beta\, \mathcal{L}_{\mathrm{dist}}(\,
                P(\mathbf{y} |\mathbf{x}, \bm{\theta}, \mathbf{M}),
                P(\mathbf{y} |\mathbf{x}, \bm{\theta}, \hat{\mathbf{M}})
            \,)}_{\text{loss modeling the agreement}}
        \end{aligned}
    $}}
\end{equation}
where $\alpha$ and $\beta$ are hyperparameters. $\mathcal{L}_{\mathrm{NLL}}$ is the conventional negative log-likelihood.
The agreement loss (i.e., $\mathcal{L}_{\mathrm{dist}}$)~\cite{Kambhatla:2022:Cipher} measures the distance between two distributions:
\begin{equation}
    \scalebox{1.0}{$
    \begin{aligned}
        \mathcal{L}_{\mathrm{dist}}(p, q) = \frac{1}{2}
        (D_{\mathrm{KL}}(p || q) + D_{\mathrm{KL}}(q || p))
    \end{aligned}
    $}
\end{equation}


\paragraph{Extension} Since our method does not change the model decoding, it can also be combined with the retrieval-based decoding algorithm as shown in $k$NN-MT~\cite{Khandelwal:2021:kNNMT}, which interpolates the model probability with a retrieved distribution. We call this decoding method $k$NN decoding.

\section{Style Customization}

\subsection{Setup}

\paragraph{NMT Model Training} In the pluggable scenario, we should first have an existing NMT model, which can serve as the foundation for further customization.
We use the training corpus of the WMT20 En$\leftrightarrow$Zh translation task\footnote{https://www.statmt.org/wmt20/translation-task.html} to train NMT models, which contains 23.9M sentence pairs. We use {\tt SentencePiece}\footnote{https://github.com/google/sentencepiece} to preprocess the data and the sentence-piece model we used is released by mBART~\cite{Liu:2020:mBART}. The architecture of our NMT models is Transformer~\cite{Vaswani:2017:Transformer}, whose hidden size is 512 and depth is 6. 
Please refer to Section~\ref{appendix:model-training} in appendix for more details.

\paragraph{Customization Data}
We evaluate the customization effect of our method in two translation directions: En$\rightarrow$Zh and Zh$\rightarrow$En. We use the works of two world-renowned writers as stylized text samples, including Shakespeare and Lu Xun. Their works created representative styles for English and Chinese, respectively. We extract their texts from the web and then split the data into training, validation, and test sets. For Shakespeare's style, the training set contains 20K English sentences while the validation and test sets contain 500 sentences, respectively. The target-language (i.e., English) training and validation sentences are then translated by the NMT model, while the test set is translated by human translators. For Lu Xun's style, the training set consists of 37K sentences while the validation and test sets contain 500 sentences. Similarly, the test set is also translated by humans while the training and validation sets are translated by NMT models. The resulting corpus is called \textbf{M}achine \textbf{T}ranslation with \textbf{S}tyle \textbf{C}ustomization (MTSC). We will add more styles in different languages for MT research in the future.

\paragraph{Memory Construction} We first build parse trees for target-side sentences using {\tt Stanford Parser}\footnote{https://nlp.stanford.edu/software/lex-parser.html} and then extract multi-granular phrases. As mentioned in Section~\ref{sec:memory-augmented-adapter}, we evenly divide the extracted phrases according to their lengths into $L$ parts to avoid information redundancy between different layers. We did not store the representations of phrases longer than a pre-specified threshold $l_{max}$, since the occurrence of long phrases is very low. $l_{max}$ is set to 10 for Zh and 8 for En.

\paragraph{Adapter Training} The general NMT model is frozen when training the memory-augmented adapter. We determine the value of the hyperparameters based on the validation performance. Specifically, the temperature in Eq.~(\ref{eq:retrieval-attention}) is set to 0.5. Both the $\alpha$ and $\beta$ in Eq.~(\ref{eq:totloss}) are set to 5. The memory dropout rate is set to $0.1$. 
We provide more details of adapter training in Section~\ref{appendix:model-training} in appendix.

\begin{table*}[t]
\setlength{\tabcolsep}{4.0pt}
    \centering
    \renewcommand{\arraystretch}{1.1}
    \caption{Automatic evaluation for style customization. We highlight the {\bf best} and \underline{second best} scores.}
    \scalebox{0.90}{
    \begin{tabular}{l | rrr | rrr | rrr }
    \specialrule{1.5pt}{-1.5pt}{0pt}
        \multirow{2}{*}{\hspace{21mm}\textbf{Method}} & \multicolumn{3}{c|}{\textbf{BLEU}$(\uparrow)$} & \multicolumn{3}{c|}{\textbf{Perplexity}$(\downarrow)$} & \multicolumn{3}{c}{\textbf{Classifier Score}$(\uparrow)$} \\ 
 & \textbf{En-Zh} & \textbf{Zh-En} & \textbf{Avg.} &\textbf{En-Zh} & \textbf{Zh-En} &\textbf{Avg.} & \textbf{En-Zh} & \textbf{Zh-En} & \textbf{Avg.} \\ 
        \specialrule{1.0pt}{-1.0pt}{0pt}
        Vanilla \cite{Vaswani:2017:Transformer} & 13.7 & 15.7 & 14.7 & 459.6 & 127.4 & 293.5 & 18.0 & 28.4 & 23.2 \\
        \cdashline{1-10}
        Extreme \cite{Michel:2018:PersonNMT} & 16.0 & 17.7 & 16.9 & 315.2 & 113.7 & 214.5 & 37.4 & 43.4 & 40.4 \\
        Adapter \cite{Houlsby:2019:Adapter} & 16.8 & 19.4 & 18.1 & 351.1 & 121.0 & 236.1 & 33.8 & 58.2 & 46.0 \\
        MT+Rewrite \cite{Syed:2020:Adapting} & 16.3 & 15.7 & 16.0 & \underline{222.4} & 127.5 & 349.8 & \underline{47.0} & 28.4 & 37.7 \\
        $k$NN-MT \cite{Khandelwal:2021:kNNMT} & 18.9 & 20.0 & 19.5 & 230.7 & \underline{98.5} & \underline{164.6} & 42.2 & \underline{70.4} & 56.3 \\
        DExperts \cite{Liu:2021:Dexperts} & 13.8 & 15.9 & 14.9 & 467.0 & 127.3 & 297.2 & 18.4 & 31.4 & 24.9 \\
        ChatGPT~\cite{chatgpt2022} & 20.0 & 13.6 & 16.8 & 620.0 & 131.8 & 375.9 & 41.4 & 24.6 & 33.0 \\
        \cdashline{1-10}
        Memory-augmented Adapter & \underline{20.8} & \underline{21.1} & \underline{21.0} & 257.8 & 110.9 & 184.3 & \underline{47.0} & 69.4 & \underline{58.2} \\
        \quad + $k$NN decoding & \textbf{21.3} & \textbf{21.8} & \textbf{21.6} & \textbf{199.6} & \textbf{95.1} & \textbf{147.4} & \textbf{53.2} & \textbf{85.2} & \textbf{69.2} \\
    \specialrule{1.5pt}{-1.5pt}{0pt}
    \end{tabular}
    }

    \label{tab:main_table}
\end{table*}

\begin{table*}[t]
\setlength{\tabcolsep}{10.0pt}
    \centering
    \renewcommand{\arraystretch}{1.1}
    \caption{Human evaluation for style customization in En$\rightarrow$Zh.
    The comparison is performed between $k$NN-MT and our method. 
    ``Win" means our method performs better. $\kappa$ denotes Fleiss' kappa.}
    \scalebox{0.90}{
    \begin{tabular}{ c | rrr | rrr | rrr }
    \specialrule{1.5pt}{-1.5pt}{0pt}
        \multirow{2}{*} {\textbf{Human}} & \multicolumn{3}{c|}{\textbf{Content Preservation}} & \multicolumn{3}{c|}{\textbf{Sentence Fluency}} & \multicolumn{3}{c}{\textbf{Style Similarity}} \\ 
  & \textbf{Win} & \textbf{Tie} & \textbf{Lose} &\textbf{Win} & \textbf{Tie} &\textbf{Lose} & \textbf{Win} & \textbf{Tie} & \textbf{Lose} \\ 
        \specialrule{1.0pt}{-1.0pt}{0pt}
        Rator 1 & 64.0\% & 12.0\% & 24.0\% & 61.0\% & 9.0\% & 30.0\% & 69.0\% & 5.0\% & 26.0\% \\
        Rator 2 & 64.0\% & 13.0\% & 23.0\% & 57.0\% & 13.0\% & 30.0\% & 63.0\% & 11.0\% & 26.0\% \\
        Rator 3 & 67.0\% & 9.0\% & 24.0\% & 62.0\% & 5.0\% & 33.0\% & 71.0\% & 6.0\% & 23.0\% \\
        \cdashline{1-10}
        Avg. & 65.0\% & 11.3\% & 23.7\% & 60.0\% & 9.0\% & 31.0\% & 67.7\% & 7.3\% & 25.0\% \\
        \hline
        $\kappa$ & \multicolumn{3}{c|}{0.476} & \multicolumn{3}{c|}{0.446} & \multicolumn{3}{c}{0.656} \\
    \specialrule{1.5pt}{-1.5pt}{0pt}
    \end{tabular}
    }

    \label{tab:human_evaluation}
\end{table*}

\paragraph{Baselines} We compare our method with the following representative baselines: 
{\em Extreme}~\cite{Michel:2018:PersonNMT}, {\em Adapter}~\cite{Houlsby:2019:Adapter}, {\em MT+Rewrite}~\cite{Syed:2020:Adapting}, {\em $k$NN-MT}~\cite{Khandelwal:2021:kNNMT}, {\em DExperts}~\cite{Liu:2021:Dexperts}, {\em ChatGPT} (GPT3.5-turbo-0301;~\citealp{chatgpt2022}).


\paragraph{Evaluation Metrics} 
We use both automatic and human evaluation to thoroughly compare them. The automatic evaluation metrics are as follows:
\begin{itemize}
    \item {\em BLEU}: measuring the translation quality of model outputs.
    We use sacreBLEU\footnote{English-Chinese: nrefs:1 $|$ case:mixed $|$ eff:no $|$ tok:zh $|$ smooth:exp $|$ version:2.3.1. Chinese-English: nrefs:1 $|$ case:mixed $|$ eff:no $|$ tok:13a $|$ smooth:exp $|$ version:2.3.1.}~\cite{Post:2018:sacreBLEU} to estimate the BLEU score.
    \item {\em Perplexity}: measuring the fluency of model outputs. We fine-tune a pretrained Transformer LM~\cite{Dai:2019:TransformerXL} with stylized text to calculate perplexity.
    \item {\em Classifier Score}: measuring the similarity between model outputs and the stylized text samples. We follow \citet{Li:2018:Delete} to train style classifiers to quantify the style similarity. The classifier we used is TextCNN~\cite{Yoon:2014:Conv}. It can achieve an accuracy of 93.5\% for Lu Xun's style and 94.5\% for Shakespeare's style. We use these classifiers to estimate whether the output is in the desired style.
\end{itemize}

\subsection{Main Results}
\paragraph{Automatic Evaluation}
Table~\ref{tab:main_table} shows the performance of all the involved methods in the style customization task.
When decoding with vanilla beam search, our method can outperform all the baselines in terms of BLEU and classifier score on average, indicating the effectiveness of the proposed memory-augmented adapter in controlling the output style of NMT models. The perplexity of $k$NN-MT is better than ours, but its BLEU score is much worse. Although ChatGPT customizes the translation at a small cost, the result is not satisfactory. When combined with $k$NN decoding, which is illustrated in {\bf Extension} in Section~\ref{sec:memory-augmented-adapter}, our method can be further improved, achieving the best performance across all the three automatic metrics. These results re-demonstrate that our method is complementary to $k$NN-MT.

\begin{table*}[t]
\setlength{\tabcolsep}{7pt}
    \centering
    \renewcommand{\arraystretch}{1.1}
    \caption{BLEU scores in the domain customization task. We highlight the {\bf best} and \underline{second best} scores.}
    \scalebox{.90}{
    \begin{tabular}{ p{4.5cm} | c c | c c | c c | c c | c c }
    \specialrule{1.5pt}{-1.5pt}{0pt}
        \multirow{2}{*} {\hspace{16mm}\textbf{Method}} & \multicolumn{2}{c|}{\textbf{IT}} & \multicolumn{2}{c|}{\textbf{Medical}} & \multicolumn{2}{c|}{\textbf{Law}} & \multicolumn{2}{c|}{\textbf{Koran}} & \multicolumn{2}{c}{\textbf{Avg.}} \\ 
 & \textbf{valid} & \textbf{test} & \textbf{valid} & \textbf{test} & \textbf{valid} & \textbf{test} & \textbf{valid} & \textbf{test} & \textbf{valid} & \textbf{test} \\ 
        \specialrule{1.0pt}{-1.0pt}{0pt}
        Vanilla & 28.1 & 28.4 & 26.4 & 27.6 & 36.2 & 35.9 & 10.9 & 11.5 & 25.4 & 25.9 \\
        \cdashline{1-11}
        Adapter & \underline{30.9} & 30.5 & 26.8 & 28.0 & 36.0 & 35.6 & 12.9 & 13.5 & 26.7 & 26.9 \\
        $k$NN-MT & 28.8 & 29.2 & \underline{30.0} & \underline{32.3} & \underline{38.3} & \underline{38.4} & 14.6 & 15.1 & 27.9 & 28.8 \\
        \cdashline{1-11}
        Memory-augmented Adapter & \textbf{31.2} & \underline{31.1} & \underline{30.0} & 32.0 & 37.5 & 37.3 & \underline{14.7} & \underline{15.3} & \underline{28.4} & \underline{28.9} \\
        \quad + $k$NN decoding & 30.5 & \textbf{31.4} & \textbf{31.3} & \textbf{33.5} & \textbf{38.8} & \textbf{38.6} & \textbf{15.7} & \textbf{16.5} & \textbf{29.1} & \textbf{30.0} \\
    \specialrule{1.5pt}{-1.5pt}{0pt}
    \end{tabular}
    }

    \label{tab:domain_adaptation}
\end{table*}

\paragraph{Human Evaluation}

We also perform a human evaluation to assess the translation quality of different methods. We follow previous works~\cite{Zhang:2018:improving,Ke:2019:araml} to ask human evaluators to compare the outputs of different methods. Since human evaluation is time-consuming and labor-intensive, we only compare our method with the strongest baseline (i.e., $k$NN-MT) in En$\rightarrow$Zh. Note that our outputs used for human evaluation are generated using vanilla beam search.
Following \citet{Hu:2022:style}, each sentence is evaluated in terms of content preservation, sentence fluency, and style similarity.
Table~\ref{tab:human_evaluation} shows the results, from which we find our approach performs better than the baseline in all the three evaluation aspects. The agreement of the three human evaluators is estimated through Fleiss' kappa~\cite{Fleiss:1971:kappa} and the results demonstrate {\em moderate agreement} $(0.4 \le \kappa \le 0.6)$ in terms of both content preservation and sentence fluency and {\em good agreement} $(0.6 \le \kappa \le 0.8)$ regarding style similarity.



\begin{figure}[t]
    \centering
    \includegraphics[width=0.42\textwidth]{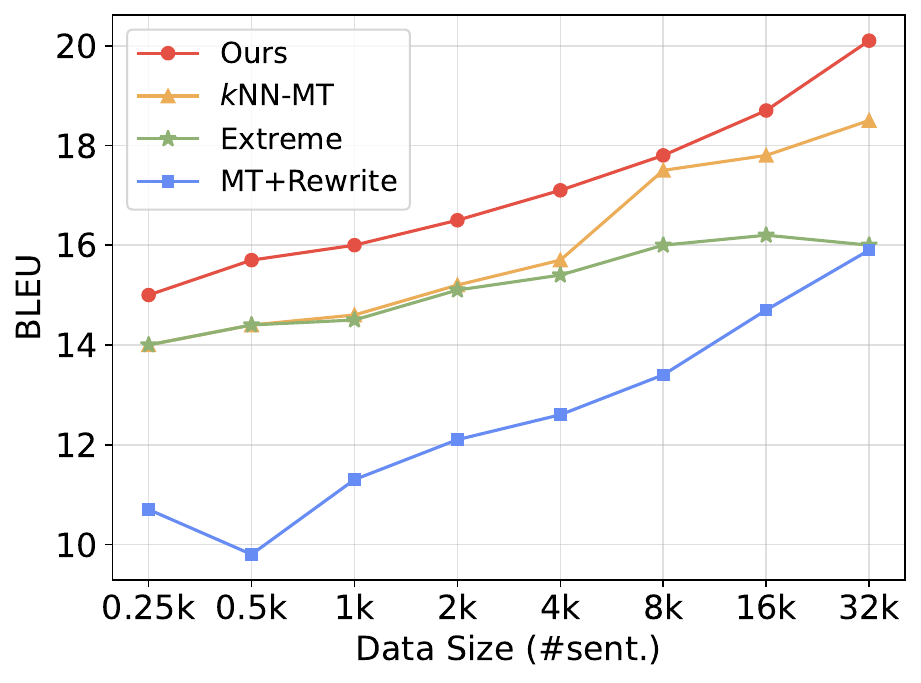}
    \caption{
        Performance of style customization at different data scales.
        ``Ours'' does not use $k$NN decoding. 
    }
\label{fig:data-scale}
\vspace{-0.8em}
\end{figure}

\subsection{Performance at Different Data Scales}

In some cases, the user-provided data can be of extremely small scale~\cite{Michel:2018:PersonNMT}. We thus investigate the performance of the involved methods using customization data of different scales. Figure~\ref{fig:data-scale} shows the results. Our memory-augmented adapter consistently outperforms the baselines at different data scales, even with only 250 exemplary sentences. These results show that our method can be applied to extremely low-resource adaptation scenarios.


\section{Domain Customization}

\subsection{Setup}

\paragraph{NMT Model Training} We train the NMT model using the WMT14 De-En training corpus\footnote{https://www.statmt.org/wmt14/translation-task.html}, including 4.5M sentence pairs. The training data is preprocessed in the same way as style customization.

\paragraph{Customization Data} To evaluate the performance in the domain customization setting, we follow previous works~\cite{Aharoni:2020:Unsupervised,Zheng:2021:Non-parametric} to use a multi-domain dataset, which includes four domains: {\em IT}, {\em Medical}, {\em Law} and {\em Koran}. To simulate real-world user customization where the user-provided data is often of small scale, we randomly select 20K sentences for IT, Medical, and Law, and use all the 18K sentences for Koran. We also use only the target-side training data to simulate real-world cases and use NMT models to generate synthetic parallel data.
All the validation and test sets are authentic parallel data.

\paragraph{Memory Construction and Adapter Training} We filter phrases longer than 10 during memory construction. For adapter training, $T$ is set to 0.1 for Medical and Law, and 0.5 for the other two domains. Both $\alpha$ and $\beta$ are set to 5 on all the four domains. The memory dropout rate is set to 0.1.

\paragraph{Baselines} We compare our proposed method with two representative pluggable domain adaptation baselines: adapter~\cite{Houlsby:2019:Adapter} and $k$NN-MT~\cite{Khandelwal:2021:kNNMT}.

\subsection{Main Results}

The adaptation performance on different domains is shown in Table~\ref{tab:domain_adaptation}. On average, our method can outperform the two baselines even without $k$NN decoding, demonstrating the effectiveness of our motivation to boost parametric plugins with external memory. When combined with $k$NN decoding, our method can achieve better results on Medical, Law, and Koran. Using $k$NN decoding, our method can improve 3.1 and 1.2 BLEU scores over Adapter and $k$NN-MT on the test sets, respectively.

\begin{figure}[t]
    \centering
    \includegraphics[width=0.45\textwidth]{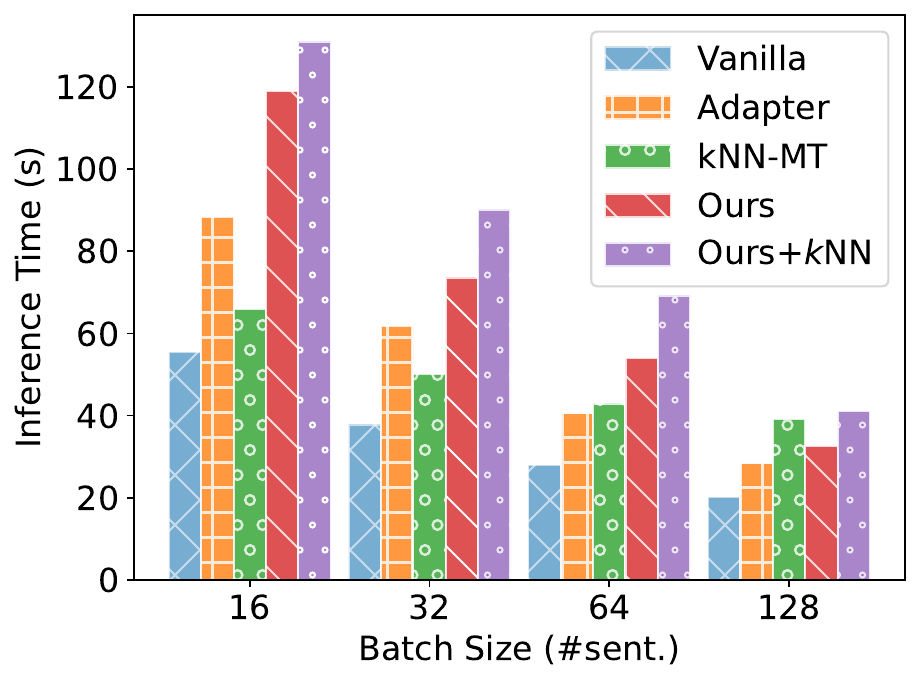}
    \caption{
        Inference time reported on the IT domain. ``Ours'' is not combined with $k$NN decoding.
    }
\label{fig:infer-time}
\end{figure}

\subsection{Inference Time}

A concern for retrieval-augmented methods is that they may significantly slow down the inference process.
As shown in Figure~\ref{fig:infer-time}, our method is slower than Adapter, but the difference between the two methods becomes very slight when using big batch sizes. For instance, our inference time is only 1.15 times that of Adapter with a batch size of 128. Our method is also comparable to $k$NN-MT. In particular, when the batch size is set to 128, our method is slightly faster than $k$NN-MT. When also using $k$NN decoding, our method is slightly slower than $k$NN-MT (i.e., 41.1s vs. 39.1s with batch size = 128). We implement the $k$NN algorithm with {\tt Faiss-gpu}~\cite{Johnson:2017:Faiss} to accelerate the retrieval process.

\section{Discussion}

\subsection{Effect of Different Components}
\label{sec:ablation-study-appendix}

We conduct thorough ablation studies to better understand the effect of the proposed components.

\begin{figure}[t]
    \centering
    \includegraphics[width=0.42\textwidth]{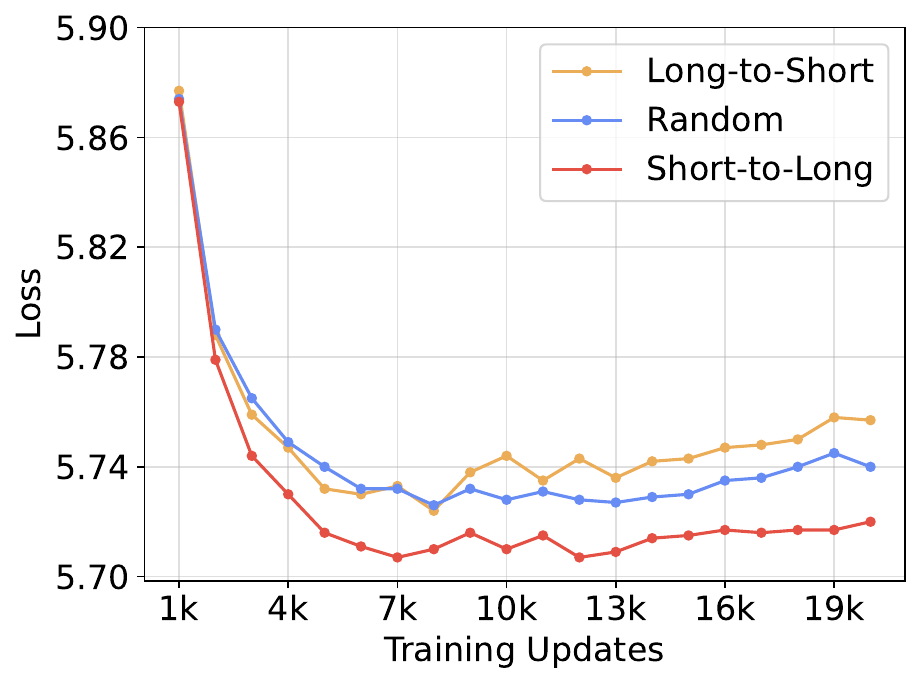}
    \caption{
        Effect of granularity distribution across the decoder layers. Each curve denotes the validation loss in the En-Zh style customization task. We only use the vanilla NLL loss to rule out the effect of the training strategy. ``Long-to-Short'': lower layers store the representations of longer phrases while higher layers store the representations of shorter layers. ``Random'': each phrase pair is stored in a randomly selected layer. ``Short-to-Long'': lower layers store shorter phrases while higher layers store longer phrases.
    }
\label{fig:multi-granu}
\end{figure}

\paragraph{Granularity Distribution} As mentioned in Section~\ref{sec:memory-construction}, we divide all the phrase pairs into several different parts to reduce the redundancy of information among the decoder layers. Our basic idea is that a certain phrase pair only needs to appear in one layer of the decoder. The phrase pairs are divided according to their lengths. We investigate three ways to distribute the phrase pairs to the decoder layers: (1) {\em long-to-short} where the phrase length decreases from the bottom layer to the top layer; (2) {\em short-to-long} where the phrase length increases from the bottom layer to the top layer; and (3) {\em random} where the memory item of a certain phrase pair is stored by a randomly selected layer. Figure~\ref{fig:multi-granu} shows the results of the three ways, where we find short-to-long achieves the best performance. We think the reason is that different layers may carry various types of linguistic properties in the Transformer model~\cite{Voita:2019:bottom-up}, which requires information of different granularities. 
When reading the memory, queries from lower layers may contain less contextualized information~\cite{Hewitt:2019:Probe}, thus short phrases are more suitable for them. At higher layers, long phrases that can provide more contextualized information performs better. We thus use short-to-long as the default setting.

\begin{table}
\renewcommand{\arraystretch}{1.2}
\caption{BLEU scores when using different memory dropout. The results are reported on the validation set in the En-Zh style customization task.}
\setlength{\tabcolsep}{15pt}
    \centering
    \scalebox{.90}{
    \begin{tabular}{ l | c }
     \specialrule{1.5pt}{-1.5pt}{0pt}
         \hspace{50pt}\textbf{Method} & \textbf{BLEU} \\ 
         \specialrule{1.0pt}{-1.0pt}{0pt}
No memory dropout & 18.2 \\
\quad + item-level memory dropout & 18.2 \\
\quad + layer-level memory dropout & 18.6 \\
    \specialrule{1.5pt}{-1.5pt}{0pt}
    \end{tabular}
    }
    \label{tab:consistent_training}
\end{table}

\paragraph{Effect of Memory Dropout}

We investigate the performance of two types of memory dropout: (1) {\em item-level memory dropout} that drop each item with a certain probability; and (2) {\em layer-level memory dropout} that drop all the memories at a decoder layer with a certain probability. Table~\ref{tab:consistent_training} shows the results, where all the models are trained using the overall loss function (i.e., $\mathcal{L}$ in Eq.~(\ref{eq:totloss})). 
We find the layer-level memory dropout performs better. Therefore, we use layer-level memory dropout by default.

\begin{figure}[t]
    \centering
    \includegraphics[width=0.42\textwidth]{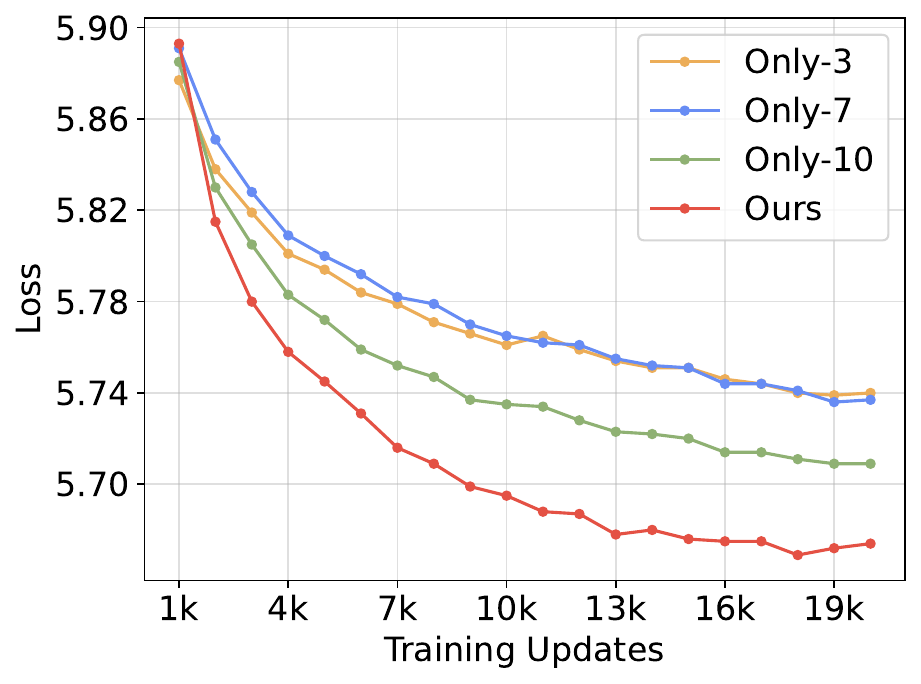}
    \caption{
        Comparison between single-granular and multi-granular. Each curve is plotted on the validation set in the En$\rightarrow$Zh style customization task. ``Only-$\mathrm{x}$'': only using phrases of length $\mathrm{x}$ to build the memory. ``Ours'': using the multi-granular memory.
    }
\label{fig:single-granu}
\end{figure}

\paragraph{Effect of Memory Granularity} To validate the necessity of building memory in a multi-granular form, we compare the performance of single- and multi-granular memories. Figure~\ref{fig:single-granu} shows the results, where we find using multi-granular memory can achieve lower validation loss, indicating the effectiveness of our method. We use multi-granular memory in other experiments by default.

\begin{table}[!t]
\renewcommand{\arraystretch}{1.2}
\caption{Analysis of memory usage.
    }
\setlength{\tabcolsep}{15pt}
    \centering
    \scalebox{.90}{
    \begin{tabular}{ l | c }
     \specialrule{1.5pt}{-1.5pt}{0pt}
         \hspace{30pt}\textbf{Method} & \textbf{BLEU} \\ 
         \specialrule{1.0pt}{-1.0pt}{0pt}
Ours & 18.6 \\
\cdashline{1-2}
\quad $-$ gated fusion & 18.3 \\
\quad $-$ source-side memory & 16.5 \\
\quad $-$ target-side memory & 16.1 \\
    \specialrule{1.5pt}{-1.5pt}{0pt}
    \end{tabular}
    }
    \label{tab:gate_datastore}
\end{table}

\paragraph{Analysis of Memory Usage} 
Table~\ref{tab:gate_datastore} shows the effect of different components that are related to memory usage. Firstly, we notice that the gated fusion mechanism has a positive effect on translation quality, indicating the necessity of learning an input-dependent interpolation ratio between original model representations and retrieved results. In addition, we observe that there is a significant performance drop when using only either the source- or target-side memory. These results demonstrate that building parallel memory using phrase pairs is very useful.

\paragraph{Effect of Integration Layers}
We also integrate the memory into different layers to better understand our method.
Table~\ref{tab:datastore_layers} shows the results, from which we find using the memories at all layers performs best and higher layers tend to be more important than lower layers. A potential reason is that memories at higher layers contain more contextualized information.

\begin{table}[!t]
\renewcommand{\arraystretch}{1.2}
\caption{Model performance when integrating memory into different layers.
    "\checkmark": equipped with the memory. "-": not equipped with the memory.
    }
\setlength{\tabcolsep}{10pt}
    \centering
    \scalebox{.90}{
    \begin{tabular}{ c c c c c c | c}
     \specialrule{1.5pt}{-1.5pt}{0pt}
         \multicolumn{6}{c|}{\textbf{Selected Layers}} & \multirow{2}{*}{\textbf{BLEU}} \\
         \textbf{L1} & \textbf{L2} & \textbf{L3} & \textbf{L4} & \textbf{L5} & \textbf{L6} & \\
         \specialrule{1.0pt}{-1.0pt}{0pt}
\checkmark & - & - & - & - & - & 15.1 \\
\checkmark & \checkmark & - & - & - & - & 15.5 \\
\checkmark & \checkmark & \checkmark & - & - & - & 15.6 \\
\checkmark & \checkmark & \checkmark & \checkmark & - & - & 16.2 \\
\checkmark & \checkmark & \checkmark & \checkmark & \checkmark & - & 16.9 \\
\checkmark & \checkmark & \checkmark & \checkmark & \checkmark & \checkmark & \textbf{18.6} \\
- & \checkmark & \checkmark & \checkmark & \checkmark & \checkmark & 17.5 \\
- & - & \checkmark & \checkmark & \checkmark & \checkmark & 17.2 \\
- & - & - & \checkmark & \checkmark & \checkmark & 17.2 \\
- & - & - & - & \checkmark & \checkmark & 17.2 \\
- & - & - & - & - & \checkmark & 16.4 \\
    \specialrule{1.5pt}{-1.5pt}{0pt}
    \end{tabular}
    }
    \label{tab:datastore_layers}
\end{table}

\begin{table}[ht]
\renewcommand{\arraystretch}{1.2}
\caption{Performance of fine-tuning all model parameters on the test set of En-Zh style customization.}
\setlength{\tabcolsep}{7pt}
    \centering
    \scalebox{.90}{
    \begin{tabular}{ l | c c c }
     \specialrule{1.5pt}{-1.5pt}{0pt}
         \hspace{25pt}\textbf{Method} & \textbf{BLEU} & \textbf{PPL} & \textbf{Class.} \\ 
         \specialrule{1.0pt}{-1.0pt}{0pt}
         Transformer & 13.7 & 459.6 & 18.0 \\
        \cdashline{1-4}
        Fine-tuning & 22.3 & 255.5 & 51.2 \\
        \ \ + Mem.-Aug. Adapter & 23.2 & 250.6 & 50.4 \\
    \specialrule{1.5pt}{-1.5pt}{0pt}
    \end{tabular}
    }
    \label{tab:compare_finetune}
\end{table}

\subsection{Comparison with Fine-tuning}

Although our goal in this work is to better build pluggable NMT models, our method is not only limited to this setting. For instance, the proposed memory-augmented adapter can also be used when the NMT model is not frozen (i.e., fine-tuning~\cite{Luong:2015:Domain}). Table~\ref{tab:compare_finetune} shows the results, where we observe that our method can also improve the performance of fine-tuning. This result implies that the external memory may provide essential information that is complementary to that stored in model parameters.



\section{Conclusion}

In this work, we propose a memory-based adapter to build pluggable NMT models, which can let the users customize the generation behavior of NMT models by simply providing some text samples. We improve both the memory design and utilization to help existing models better adapt to the user-demanded styles or domains. Experiments demonstrate the superiority of our proposed method over several representative baselines. By changing the memory format, we believe our method can be applied to some other sequence generation tasks.


\section*{Acknowledgements}

This work is supported by the National Key R\&D Program of China (2022ZD0160502) and the National Natural Science Foundation of China (No. 61925601, 62276152, 62236011). We are also grateful to Ziang Liu and Yourun Lin for their dedicated efforts in the human evaluation process, which greatly enhanced the quality this study.

\nocite{*}
\section{Bibliographical References}
\label{sec:reference}
\vspace{-1.5em}

\bibliographystyle{lrec-coling2024-natbib}
\bibliography{lrec-coling2024-example}

\begin{thebibliography}{71}
\expandafter\ifx\csname natexlab\endcsname\relax\def\natexlab#1{#1}\fi

\bibitem[{Aharoni and Goldberg(2020)}]{Aharoni:2020:Unsupervised}
Roee Aharoni and Yoav Goldberg. 2020.
\newblock \href {https://doi.org/10.18653/v1/2020.acl-main.692} {Unsupervised domain clusters in pretrained language models}.
\newblock In \emph{Proceedings of ACL 2020}, pages 7747--7763.

\bibitem[{Bahdanau et~al.(2015)Bahdanau, Cho, and Bengio}]{Bahdanau:2015:RNNSearch}
Dzmitry Bahdanau, Kyunghyun Cho, and Yoshua Bengio. 2015.
\newblock \href {https://arxiv.org/abs/1409.0473} {Neural machine translation by jointly learning to align and translate}.
\newblock In \emph{Proceedings of ICLR 2015}.

\bibitem[{Bapna and Firat(2019)}]{Bapna:2019:Adapter}
Ankur Bapna and Orhan Firat. 2019.
\newblock \href {https://doi.org/10.18653/v1/D19-1165} {Simple, scalable adaptation for neural machine translation}.
\newblock In \emph{Proceedings of EMNLP-IJCNLP 2019}, pages 1538--1548.

\bibitem[{Barrault et~al.(2020)Barrault, Biesialska, Bojar, Costa-juss{\`a}, Federmann, Graham, Grundkiewicz, Haddow, Huck, Joanis, Kocmi, Koehn, Lo, Ljube{\v{s}}i{\'c}, Monz, Morishita, Nagata, Nakazawa, Pal, Post, and Zampieri}]{Barrault:2020:WMT}
Lo{\"\i}c Barrault, Magdalena Biesialska, Ond{\v{r}}ej Bojar, Marta~R. Costa-juss{\`a}, Christian Federmann, Yvette Graham, Roman Grundkiewicz, Barry Haddow, Matthias Huck, Eric Joanis, Tom Kocmi, Philipp Koehn, Chi-kiu Lo, Nikola Ljube{\v{s}}i{\'c}, Christof Monz, Makoto Morishita, Masaaki Nagata, Toshiaki Nakazawa, Santanu Pal, Matt Post, and Marcos Zampieri. 2020.
\newblock \href {https://aclanthology.org/2020.wmt-1.1} {Findings of the 2020 conference on machine translation ({WMT}20)}.
\newblock In \emph{Proceedings of the 5th Conference on Machine Translation}, pages 1--55.

\bibitem[{Bojar et~al.(2014)Bojar, Buck, Federmann, Haddow, Koehn, Leveling, Monz, Pecina, Post, Saint-Amand, Soricut, Specia, and Tamchyna}]{Bojar:2014:WMT}
Ond{\v{r}}ej Bojar, Christian Buck, Christian Federmann, Barry Haddow, Philipp Koehn, Johannes Leveling, Christof Monz, Pavel Pecina, Matt Post, Herve Saint-Amand, Radu Soricut, Lucia Specia, and Ale{\v{s}} Tamchyna. 2014.
\newblock \href {https://doi.org/10.3115/v1/W14-3302} {Findings of the 2014 workshop on statistical machine translation}.
\newblock In \emph{Proceedings of the 9th Workshop on Statistical Machine Translation}, pages 12--58.

\bibitem[{Borgeaud et~al.(2022)Borgeaud, Mensch, Hoffmann, Cai, Rutherford, Millican, Van Den~Driessche, Lespiau, Damoc, Clark, De~Las~Casas, Guy, Menick, Ring, Hennigan, Huang, Maggiore, Jones, Cassirer, Brock, Paganini, Irving, Vinyals, Osindero, Simonyan, Rae, Elsen, and Sifre}]{Chaudhuri:2022:Retro}
Sebastian Borgeaud, Arthur Mensch, Jordan Hoffmann, Trevor Cai, Eliza Rutherford, Katie Millican, George~Bm Van Den~Driessche, Jean-Baptiste Lespiau, Bogdan Damoc, Aidan Clark, Diego De~Las~Casas, Aurelia Guy, Jacob Menick, Roman Ring, Tom Hennigan, Saffron Huang, Loren Maggiore, Chris Jones, Albin Cassirer, Andy Brock, Michela Paganini, Geoffrey Irving, Oriol Vinyals, Simon Osindero, Karen Simonyan, Jack Rae, Erich Elsen, and Laurent Sifre. 2022.
\newblock \href {https://proceedings.mlr.press/v162/borgeaud22a.html} {Improving language models by retrieving from trillions of tokens}.
\newblock In \emph{Proceedings of ICML 2022}, pages 2206--2240.

\bibitem[{Chen et~al.(2021)Chen, Sun, and Liu}]{Chen:2021:maskalign}
Chi Chen, Maosong Sun, and Yang Liu. 2021.
\newblock \href {https://doi.org/10.18653/v1/2021.acl-long.369} {Mask-align: Self-supervised neural word alignment}.
\newblock In \emph{Proceedings of ACL-IJCNLP 2021}, pages 4781--4791.

\bibitem[{Chen et~al.(2022)Chen, Verga, de~Jong, Wieting, and Cohen}]{Chen:2022:Augmenting}
Wenhu Chen, Pat Verga, Michiel de~Jong, John Wieting, and William Cohen. 2022.
\newblock Augmenting pre-trained language models with qa-memory for open-domain question answering.
\newblock \emph{arXiv preprint arXiv:2204.04581}.

\bibitem[{Chu and Wang(2018)}]{Chu:2018:Domain}
Chenhui Chu and Rui Wang. 2018.
\newblock \href {https://www.aclweb.org/anthology/C18-1111} {A survey of domain adaptation for neural machine translation}.
\newblock In \emph{Proceedings of COLING 2018}, pages 1304--1319.

\bibitem[{Dai et~al.(2019)Dai, Yang, Yang, Carbonell, Le, and Salakhutdinov}]{Dai:2019:TransformerXL}
Zihang Dai, Zhilin Yang, Yiming Yang, Jaime~G Carbonell, Quoc Le, and Ruslan Salakhutdinov. 2019.
\newblock Transformer-{X}{L}: Attentive language models beyond a fixed-length context.
\newblock In \emph{Proceedings of ACL 2019}, pages 2978--2988.

\bibitem[{Dathathri et~al.(2020)Dathathri, Madotto, Lan, Hung, Frank, Molino, Yosinski, and Liu}]{Dathathri:2020:PPLM}
Sumanth Dathathri, Andrea Madotto, Janice Lan, Jane Hung, Eric Frank, Piero Molino, Jason Yosinski, and Rosanne Liu. 2020.
\newblock \href {https://openreview.net/forum?id=H1edEyBKDS} {Plug and play language models: A simple approach to controlled text generation}.
\newblock In \emph{Proceedings of ICLR 2020}.

\bibitem[{de~Jong et~al.(2022)de~Jong, Zemlyanskiy, FitzGerald, Sha, and Cohen}]{Jong:2022:Mention}
Michiel de~Jong, Yury Zemlyanskiy, Nicholas FitzGerald, Fei Sha, and William~W. Cohen. 2022.
\newblock \href {https://openreview.net/forum?id=OY1A8ejQgEX} {Mention memory: incorporating textual knowledge into transformers through entity mention attention}.
\newblock In \emph{Proceedings of ICLR 2022}.

\bibitem[{Ding et~al.(2022)Ding, Qin, Yang, Wei, Yang, Su, Hu, Chen, Chan, Chen, Yi, Zhao, Wang, Liu, Zheng, Chen, Liu, Tang, Li, and Sun}]{Ding:2022:DeltaTuning}
Ning Ding, Yujia Qin, Guang Yang, Fuchao Wei, Zonghan Yang, Yusheng Su, Shengding Hu, Yulin Chen, Chi-Min Chan, Weize Chen, Jing Yi, Weilin Zhao, Xiaozhi Wang, Zhiyuan Liu, Hai-Tao Zheng, Jianfei Chen, Yang Liu, Jie Tang, Juanzi Li, and Maosong Sun. 2022.
\newblock \href {https://arxiv.org/abs/2203.06904} {Delta tuning: A comprehensive study of parameter efficient methods for pre-trained language models}.
\newblock \emph{arXiv preprint arXiv:2203.06904}.

\bibitem[{Dyer et~al.(2013)Dyer, Chahuneau, and Smith}]{Dyer:2013:fastalign}
Chris Dyer, Victor Chahuneau, and Noah~A. Smith. 2013.
\newblock \href {https://aclanthology.org/N13-1073} {A simple, fast, and effective reparameterization of {IBM} model 2}.
\newblock In \emph{Proceedings of NAACL 2013}, pages 644--648.

\bibitem[{Fan et~al.(2021)Fan, Bhosale, Schwenk, Ma, El-Kishky, Goyal, Baines, Celebi, Wenzek, Chaudhary, Goyal, Birch, Liptchinsky, Edunov, Auli, and Joulin}]{Fan:2021:m2m100}
Angela Fan, Shruti Bhosale, Holger Schwenk, Zhiyi Ma, Ahmed El-Kishky, Siddharth Goyal, Mandeep Baines, Onur Celebi, Guillaume Wenzek, Vishrav Chaudhary, Naman Goyal, Tom Birch, Vitaliy Liptchinsky, Sergey Edunov, Michael Auli, and Armand Joulin. 2021.
\newblock \href {http://jmlr.org/papers/v22/20-1307.html} {Beyond english-centric multilingual machine translation}.
\newblock \emph{Journal of Machine Learning Research}, 22(107):1--48.

\bibitem[{Fleiss(1971)}]{Fleiss:1971:kappa}
Joseph~L Fleiss. 1971.
\newblock Measuring nominal scale agreement among many raters.
\newblock \emph{Psychological bulletin}, 76(5):378.

\bibitem[{Hassan et~al.(2018)Hassan, Aue, Chen, Chowdhary, Clark, Federmann, Huang, Junczys{-}Dowmunt, Lewis, Li, Liu, Liu, Luo, Menezes, Qin, Seide, Tan, Tian, Wu, Wu, Xia, Zhang, Zhang, and Zhou}]{Hassan:2018:Human}
Hany Hassan, Anthony Aue, Chang Chen, Vishal Chowdhary, Jonathan Clark, Christian Federmann, Xuedong Huang, Marcin Junczys{-}Dowmunt, William Lewis, Mu~Li, Shujie Liu, Tie{-}Yan Liu, Renqian Luo, Arul Menezes, Tao Qin, Frank Seide, Xu~Tan, Fei Tian, Lijun Wu, Shuangzhi Wu, Yingce Xia, Dongdong Zhang, Zhirui Zhang, and Ming Zhou. 2018.
\newblock \href {http://arxiv.org/abs/1803.05567} {Achieving human parity on automatic chinese to english news translation}.
\newblock \emph{CoRR}, abs/1803.05567.

\bibitem[{He et~al.(2021{\natexlab{a}})He, Neubig, and Berg-Kirkpatrick}]{He:2021:EfficientKNN}
Junxian He, Graham Neubig, and Taylor Berg-Kirkpatrick. 2021{\natexlab{a}}.
\newblock \href {https://doi.org/10.18653/v1/2021.emnlp-main.461} {Efficient nearest neighbor language models}.
\newblock In \emph{Proceedings of EMNLP 2021}, pages 5703--5714.

\bibitem[{He et~al.(2021{\natexlab{b}})He, Huang, Cui, Li, and Liu}]{He:2021:Fast}
Qiuxiang He, Guoping Huang, Qu~Cui, Li~Li, and Lemao Liu. 2021{\natexlab{b}}.
\newblock Fast and accurate neural machine translation with translation memory.
\newblock In \emph{Proceedings of ACL-IJCNLP 2021}, pages 3170--3180.

\bibitem[{Hewitt and Liang(2019)}]{Hewitt:2019:Probe}
John Hewitt and Percy Liang. 2019.
\newblock \href {https://doi.org/10.18653/v1/D19-1275} {Designing and interpreting probes with control tasks}.
\newblock In \emph{Proceedings of EMNLP-IJCNLP 2019}, pages 2733--2743.

\bibitem[{Hochreiter and Schmidhuber(1997)}]{hochreiter1997long}
Sepp Hochreiter and J{\"u}rgen Schmidhuber. 1997.
\newblock Long short-term memory.
\newblock \emph{Neural computation}, 9(8):1735--1780.

\bibitem[{Houlsby et~al.(2019)Houlsby, Giurgiu, Jastrzebski, Morrone, De~Laroussilhe, Gesmundo, Attariyan, and Gelly}]{Houlsby:2019:Adapter}
Neil Houlsby, Andrei Giurgiu, Stanislaw Jastrzebski, Bruna Morrone, Quentin De~Laroussilhe, Andrea Gesmundo, Mona Attariyan, and Sylvain Gelly. 2019.
\newblock \href {https://proceedings.mlr.press/v97/houlsby19a.html} {Parameter-efficient transfer learning for {NLP}}.
\newblock In \emph{Proceedings of ICML 2019}, pages 2790--2799.

\bibitem[{Hu et~al.(2022{\natexlab{a}})Hu, yelong shen, Wallis, Allen-Zhu, Li, Wang, Wang, and Chen}]{Hu:2022:LoRA}
Edward~J Hu, yelong shen, Phillip Wallis, Zeyuan Allen-Zhu, Yuanzhi Li, Shean Wang, Lu~Wang, and Weizhu Chen. 2022{\natexlab{a}}.
\newblock \href {https://openreview.net/forum?id=nZeVKeeFYf9} {Lo{RA}: Low-rank adaptation of large language models}.
\newblock In \emph{Proceedings of ICLR 2022}.

\bibitem[{Hu et~al.(2019)Hu, Xia, Neubig, and Carbonell}]{Hu:2019:Domain}
Junjie Hu, Mengzhou Xia, Graham Neubig, and Jaime Carbonell. 2019.
\newblock \href {https://doi.org/10.18653/v1/P19-1286} {Domain adaptation of neural machine translation by lexicon induction}.
\newblock In \emph{Proceedings of ACL 2019}, pages 2989--3001.

\bibitem[{Hu et~al.(2022{\natexlab{b}})Hu, Lee, Aggarwal, and Zhang}]{Hu:2022:style}
Zhiqiang Hu, Roy Ka-Wei Lee, Charu~C Aggarwal, and Aston Zhang. 2022{\natexlab{b}}.
\newblock Text style transfer: A review and experimental evaluation.
\newblock \emph{ACM SIGKDD Explorations Newsletter}, 24(1):14--45.

\bibitem[{Jing et~al.(2022)Jing, Mao, Yang, Zhan, Song, Wang, and Tao}]{jing2022learning}
Yongcheng Jing, Yining Mao, Yiding Yang, Yibing Zhan, Mingli Song, Xinchao Wang, and Dacheng Tao. 2022.
\newblock Learning graph neural networks for image style transfer.
\newblock In \emph{ECCV 2022}, pages 111--128. Springer.

\bibitem[{Johnson et~al.(2017)Johnson, Douze, and J{\'e}gou}]{Johnson:2017:Faiss}
Jeff Johnson, Matthijs Douze, and Herv{\'e} J{\'e}gou. 2017.
\newblock Billion-scale similarity search with gpus.
\newblock \emph{IEEE Transactions on Big Data}, 7:535--547.

\bibitem[{Kambhatla et~al.(2022)Kambhatla, Born, and Sarkar}]{Kambhatla:2022:Cipher}
Nishant Kambhatla, Logan Born, and Anoop Sarkar. 2022.
\newblock \href {https://doi.org/10.18653/v1/2022.acl-long.17} {{C}ipher{DA}ug: Ciphertext based data augmentation for neural machine translation}.
\newblock In \emph{Proceedings of ACL 2022}, pages 201--218.

\bibitem[{Ke et~al.(2019)Ke, Huang, Huang, and Zhu}]{Ke:2019:araml}
Pei Ke, Fei Huang, Minlie Huang, and Xiaoyan Zhu. 2019.
\newblock \href {https://doi.org/10.18653/v1/D19-1436} {{ARAML}: A stable adversarial training framework for text generation}.
\newblock In \emph{Proceedings of EMNLP-IJCNLP 2019}, pages 4271--4281.

\bibitem[{Keskar et~al.(2019)Keskar, McCann, Varshney, Xiong, and Socher}]{Keskar:2019:CTRL}
Nitish~Shirish Keskar, Bryan McCann, Lav Varshney, Caiming Xiong, and Richard Socher. 2019.
\newblock {CTRL - A Conditional Transformer Language Model for Controllable Generation}.
\newblock \emph{arXiv preprint arXiv:1909.05858}.

\bibitem[{Khandelwal et~al.(2021)Khandelwal, Fan, Jurafsky, Zettlemoyer, and Lewis}]{Khandelwal:2021:kNNMT}
Urvashi Khandelwal, Angela Fan, Dan Jurafsky, Luke Zettlemoyer, and Mike Lewis. 2021.
\newblock \href {https://openreview.net/forum?id=7wCBOfJ8hJM} {Nearest neighbor machine translation}.
\newblock In \emph{Proceedings of ICLR 2021}.

\bibitem[{Khandelwal et~al.(2020)Khandelwal, Levy, Jurafsky, Zettlemoyer, and Lewis}]{Khandelwal:2020:kNNLM}
Urvashi Khandelwal, Omer Levy, Dan Jurafsky, Luke Zettlemoyer, and Mike Lewis. 2020.
\newblock \href {https://openreview.net/forum?id=HklBjCEKvH} {Generalization through memorization: Nearest neighbor language models}.
\newblock In \emph{Proceedings of ICLR 2020}.

\bibitem[{Kim(2014)}]{Yoon:2014:Conv}
Yoon Kim. 2014.
\newblock \href {https://doi.org/10.3115/v1/D14-1181} {Convolutional neural networks for sentence classification}.
\newblock In \emph{Proceedings of EMNLP 2014}, pages 1746--1751.

\bibitem[{Kingma and Ba(2014)}]{kingma2014adam}
Diederik~P Kingma and Jimmy Ba. 2014.
\newblock Adam: A method for stochastic optimization.
\newblock \emph{arXiv preprint arXiv:1412.6980}.

\bibitem[{Kirkpatrick et~al.(2017)Kirkpatrick, Pascanu, Rabinowitz, Veness, Desjardins, Rusu, Milan, Quan, Ramalho, Grabska-Barwinska et~al.}]{Kirkpatrick:2017:CF}
James Kirkpatrick, Razvan Pascanu, Neil Rabinowitz, Joel Veness, Guillaume Desjardins, Andrei~A Rusu, Kieran Milan, John Quan, Tiago Ramalho, Agnieszka Grabska-Barwinska, et~al. 2017.
\newblock Overcoming catastrophic forgetting in neural networks.
\newblock \emph{Proceedings of the national academy of sciences}, 114(13):3521--3526.

\bibitem[{Kocmi et~al.(2022)Kocmi, Bawden, Bojar, Dvorkovich, Federmann, Fishel, Gowda, Graham, Grundkiewicz, Haddow, Knowles, Koehn, Monz, Morishita, Nagata, Nakazawa, Nov\'{a}k, Popel, Popovi\'{c}, and Shmatova}]{Kocmi:2022:WMT}
Tom Kocmi, Rachel Bawden, Ond\v{r}ej Bojar, Anton Dvorkovich, Christian Federmann, Mark Fishel, Thamme Gowda, Yvette Graham, Roman Grundkiewicz, Barry Haddow, Rebecca Knowles, Philipp Koehn, Christof Monz, Makoto Morishita, Masaaki Nagata, Toshiaki Nakazawa, Michal Nov\'{a}k, Martin Popel, Maja Popovi\'{c}, and Mariya Shmatova. 2022.
\newblock \href {https://aclanthology.org/2022.wmt-1.1} {Findings of the 2022 conference on machine translation ({WMT}22)}.
\newblock In \emph{Proceedings of the 7th Conference on Machine Translation}, pages 1--45.

\bibitem[{Lewis et~al.(2020)Lewis, Perez, Piktus, Petroni, Karpukhin, Goyal, K\"{u}ttler, Lewis, Yih, Rockt\"{a}schel, Riedel, and Kiela}]{Lewis:2020:Retrieve}
Patrick Lewis, Ethan Perez, Aleksandra Piktus, Fabio Petroni, Vladimir Karpukhin, Naman Goyal, Heinrich K\"{u}ttler, Mike Lewis, Wen-tau Yih, Tim Rockt\"{a}schel, Sebastian Riedel, and Douwe Kiela. 2020.
\newblock Retrieval-augmented generation for knowledge-intensive nlp tasks.
\newblock In \emph{Advances of NeurIPS 2020}, pages 9459--9474.

\bibitem[{Li et~al.(2022)Li, Zheng, Jing, Jiao, Xiao, and Zhu}]{Li:2022:multiscale}
Bei Li, Tong Zheng, Yi~Jing, Chengbo Jiao, Tong Xiao, and Jingbo Zhu. 2022.
\newblock \href {https://proceedings.mlr.press/v162/li22ac.html} {Learning multiscale transformer models for sequence generation}.
\newblock In \emph{Proceedings of ICML 2022}, pages 13225--13241.

\bibitem[{Li et~al.(2018)Li, Jia, He, and Liang}]{Li:2018:Delete}
Juncen Li, Robin Jia, He~He, and Percy Liang. 2018.
\newblock Delete, retrieve, generate: a simple approach to sentiment and style transfer.
\newblock In \emph{Proceedings of NAACL 2018}, pages 1865--1874.

\bibitem[{Li and Liang(2021)}]{Li:2021:Prefix}
Xiang~Lisa Li and Percy Liang. 2021.
\newblock \href {https://doi.org/10.18653/v1/2021.acl-long.353} {Prefix-tuning: Optimizing continuous prompts for generation}.
\newblock In \emph{Proceedings of ACL 2021}, pages 4582--4597.

\bibitem[{Liu et~al.(2021)Liu, Sap, Lu, Swayamdipta, Bhagavatula, Smith, and Choi}]{Liu:2021:Dexperts}
Alisa Liu, Maarten Sap, Ximing Lu, Swabha Swayamdipta, Chandra Bhagavatula, Noah~A. Smith, and Yejin Choi. 2021.
\newblock \href {https://doi.org/10.18653/v1/2021.acl-long.522} {{DE}xperts: Decoding-time controlled text generation with experts and anti-experts}.
\newblock In \emph{Proceedings of ACL-IJCNLP 2021}, pages 6691--6706.

\bibitem[{Liu et~al.(2020)Liu, Gu, Goyal, Li, Edunov, Ghazvininejad, Lewis, and Zettlemoyer}]{Liu:2020:mBART}
Yinhan Liu, Jiatao Gu, Naman Goyal, Xian Li, Sergey Edunov, Marjan Ghazvininejad, Mike Lewis, and Luke Zettlemoyer. 2020.
\newblock \href {https://doi.org/10.1162/tacl_a_00343} {Multilingual denoising pre-training for neural machine translation}.
\newblock \emph{Transactions of the ACL}, 8:726--742.

\bibitem[{Luong and Manning(2015)}]{Luong:2015:Domain}
Minh-Thang Luong and Christopher Manning. 2015.
\newblock \href {https://aclanthology.org/2015.iwslt-evaluation.11} {{S}tanford neural machine translation systems for spoken language domains}.
\newblock In \emph{Proceedings of the 12th IWSLT: Evaluation Campaign}, pages 76--79.

\bibitem[{Mao et~al.(2022)Mao, Mathias, Hou, Almahairi, Ma, Han, Yih, and Khabsa}]{Mao:2022:Unipelt}
Yuning Mao, Lambert Mathias, Rui Hou, Amjad Almahairi, Hao Ma, Jiawei Han, Scott Yih, and Madian Khabsa. 2022.
\newblock \href {https://doi.org/10.18653/v1/2022.acl-long.433} {{U}ni{PELT}: A unified framework for parameter-efficient language model tuning}.
\newblock In \emph{Proceedings of ACL 2022}, pages 6253--6264.

\bibitem[{Michel and Neubig(2018)}]{Michel:2018:PersonNMT}
Paul Michel and Graham Neubig. 2018.
\newblock \href {https://doi.org/10.18653/v1/P18-2050} {Extreme adaptation for personalized neural machine translation}.
\newblock In \emph{Proceedings of ACL 2018}, pages 312--318.

\bibitem[{Niu and Carpuat(2020)}]{Niu:2020:Style}
Xing Niu and Marine Carpuat. 2020.
\newblock \href {https://aaai.org/ojs/index.php/AAAI/article/view/6379} {Controlling neural machine translation formality with synthetic supervision}.
\newblock In \emph{Proceedings of AAAI 2020}, pages 8568--8575.

\bibitem[{Niu et~al.(2017)Niu, Martindale, and Carpuat}]{Niu:2017:Style}
Xing Niu, Marianna Martindale, and Marine Carpuat. 2017.
\newblock \href {https://doi.org/10.18653/v1/D17-1299} {A study of style in machine translation: Controlling the formality of machine translation output}.
\newblock In \emph{Proceedings of EMNLP 2017}, pages 2814--2819.

\bibitem[{Niu et~al.(2018)Niu, Rao, and Carpuat}]{Niu:2018:Style}
Xing Niu, Sudha Rao, and Marine Carpuat. 2018.
\newblock \href {https://aclanthology.org/C18-1086} {Multi-task neural models for translating between styles within and across languages}.
\newblock In \emph{Proceedings of COLING 2018}, pages 1008--1021.

\bibitem[{{NLLB Team} et~al.(2022){NLLB Team}, Costa-juss{\`a}, Cross, {\c{C}}elebi, Elbayad, Heafield, Heffernan, Kalbassi, Lam, Licht, Maillard, Sun, Wang, Wenzek, Youngblood, Akula, Barrault, Gonzalez, Hansanti, Hoffman, Jarrett, Sadagopan, Rowe, Spruit, Tran, Andrews, Ayan, Bhosale, Edunov, Fan, Gao, Goswami, Guzm{\'a}n, Koehn, Mourachko, Ropers, Saleem, Schwenk, and Wang}]{NLLB:2022:NLLB}
{NLLB Team}, Marta~R. Costa-juss{\`a}, James Cross, Onur {\c{C}}elebi, Maha Elbayad, Kenneth Heafield, Kevin Heffernan, Elahe Kalbassi, Janice Lam, Daniel Licht, Jean Maillard, Anna Sun, Skyler Wang, Guillaume Wenzek, Al~Youngblood, Bapi Akula, Loic Barrault, Gabriel~Mejia Gonzalez, Prangthip Hansanti, John Hoffman, Semarley Jarrett, Kaushik~Ram Sadagopan, Dirk Rowe, Shannon Spruit, Chau Tran, Pierre Andrews, Necip~Fazil Ayan, Shruti Bhosale, Sergey Edunov, Angela Fan, Cynthia Gao, Vedanuj Goswami, Francisco Guzm{\'a}n, Philipp Koehn, Alexandre Mourachko, Christophe Ropers, Safiyyah Saleem, Holger Schwenk, and Jeff Wang. 2022.
\newblock \href {https://arxiv.org/abs/2207.04672} {No language left behind: Scaling human-centered machine translation}.
\newblock \emph{arXiv preprint arXiv:2207.04672}.

\bibitem[{OpenAI(2022)}]{chatgpt2022}
OpenAI. 2022.
\newblock \href {https://openai.com/blog/chatgpt} {Introducing {ChatGPT}}.
\newblock (Accessed on Jun 18, 2023).

\bibitem[{Papineni et~al.(2002)Papineni, Roukos, Ward, and Zhu}]{papineni2002bleu}
Kishore Papineni, Salim Roukos, Todd Ward, and Wei-Jing Zhu. 2002.
\newblock Bleu: a method for automatic evaluation of machine translation.
\newblock In \emph{Proceedings of ACL 2002}, pages 311--318.

\bibitem[{Petroni et~al.(2019)Petroni, Rockt{\"a}schel, Riedel, Lewis, Bakhtin, Wu, and Miller}]{Petroni:2019:LMKB}
Fabio Petroni, Tim Rockt{\"a}schel, Sebastian Riedel, Patrick Lewis, Anton Bakhtin, Yuxiang Wu, and Alexander Miller. 2019.
\newblock \href {https://doi.org/10.18653/v1/D19-1250} {Language models as knowledge bases?}
\newblock In \emph{Proceedings of EMNLP-IJCNLP}, pages 2463--2473.

\bibitem[{Pfeiffer et~al.(2021)Pfeiffer, Kamath, R{\"u}ckl{\'e}, Cho, and Gurevych}]{Pfeiffer:2021:Adapterfusion}
Jonas Pfeiffer, Aishwarya Kamath, Andreas R{\"u}ckl{\'e}, Kyunghyun Cho, and Iryna Gurevych. 2021.
\newblock \href {https://doi.org/10.18653/v1/2021.eacl-main.39} {{A}dapter{F}usion: Non-destructive task composition for transfer learning}.
\newblock In \emph{Proceedings of EACL 2021}, pages 487--503.

\bibitem[{Post(2018)}]{Post:2018:sacreBLEU}
Matt Post. 2018.
\newblock A call for clarity in reporting bleu scores.
\newblock In \emph{Proceedings of the 3rd Conference on Machine Translation: Research Papers}, pages 186--191.

\bibitem[{R{\"u}ckl{\'e} et~al.(2021)R{\"u}ckl{\'e}, Geigle, Glockner, Beck, Pfeiffer, Reimers, and Gurevych}]{Ruckle:2021:Adapterdrop}
Andreas R{\"u}ckl{\'e}, Gregor Geigle, Max Glockner, Tilman Beck, Jonas Pfeiffer, Nils Reimers, and Iryna Gurevych. 2021.
\newblock \href {https://doi.org/10.18653/v1/2021.emnlp-main.626} {{AdapterDrop}: {O}n the efficiency of adapters in transformers}.
\newblock In \emph{Proceedings of EMNLP 2021}, pages 7930--7946.

\bibitem[{Ruiz et~al.(2023)Ruiz, Li, Jampani, Pritch, Rubinstein, and Aberman}]{ruiz2023dreambooth}
Nataniel Ruiz, Yuanzhen Li, Varun Jampani, Yael Pritch, Michael Rubinstein, and Kfir Aberman. 2023.
\newblock Dreambooth: Fine tuning text-to-image diffusion models for subject-driven generation.
\newblock In \emph{Proceedings of CVPR 2023}, pages 22500--22510.

\bibitem[{Singh et~al.(2021)Singh, Verma, Garimella, and Srinivasan}]{singh2021drag}
Hrituraj Singh, Gaurav Verma, Aparna Garimella, and Balaji~Vasan Srinivasan. 2021.
\newblock {DRAG}: Director-generator language modelling framework for non-parallel author stylized rewriting.
\newblock In \emph{Proceedings of EACL 2021}, pages 863--873.

\bibitem[{Srivastava et~al.(2014)Srivastava, Hinton, Krizhevsky, Sutskever, and Salakhutdinov}]{Srivastava:2014:Dropout}
Nitish Srivastava, Geoffrey Hinton, Alex Krizhevsky, Ilya Sutskever, and Ruslan Salakhutdinov. 2014.
\newblock \href {http://jmlr.org/papers/v15/srivastava14a.html} {Dropout: A simple way to prevent neural networks from overfitting}.
\newblock \emph{Journal of Machine Learning Research}, 15(56):1929--1958.

\bibitem[{Syed et~al.(2020)Syed, Verma, Srinivasan, Natarajan, and Varma}]{Syed:2020:Adapting}
Bakhtiyar Syed, Gaurav Verma, Balaji~Vasan Srinivasan, Anandhavelu Natarajan, and Vasudeva Varma. 2020.
\newblock Adapting language models for non-parallel author-stylized rewriting.
\newblock In \emph{Proceedings of AAAI 2020}, pages 9008--9015.

\bibitem[{Vaswani et~al.(2017)Vaswani, Shazeer, Parmar, Uszkoreit, Jones, Gomez, Kaiser, and Polosukhin}]{Vaswani:2017:Transformer}
Ashish Vaswani, Noam Shazeer, Niki Parmar, Jakob Uszkoreit, Llion Jones, Aidan~N Gomez, {\L}ukasz Kaiser, and Illia Polosukhin. 2017.
\newblock \href {https://papers.nips.cc/paper/2017/file/3f5ee243547dee91fbd053c1c4a845aa-Paper.pdf} {Attention is all you need}.
\newblock In \emph{Advances of NeurIPS 2017}, pages 5998--6008.

\bibitem[{Voita et~al.(2019)Voita, Sennrich, and Titov}]{Voita:2019:bottom-up}
Elena Voita, Rico Sennrich, and Ivan Titov. 2019.
\newblock \href {https://doi.org/10.18653/v1/D19-1448} {The bottom-up evolution of representations in the transformer: A study with machine translation and language modeling objectives}.
\newblock In \emph{Proceedings of EMNLP-IJCNLP 2019}, pages 4396--4406.

\bibitem[{Vu and Moschitti(2021)}]{Vu:2021:Customize}
Thuy Vu and Alessandro Moschitti. 2021.
\newblock \href {http://arxiv.org/abs/2102.10243} {Machine translation customization via automatic training data selection from the web}.
\newblock \emph{CoRR}, abs/2102.10243.

\bibitem[{Wang et~al.(2021)Wang, Tu, Tan, Shi, Sun, and Liu}]{Wang:2021:LCB}
Shuo Wang, Zhaopeng Tu, Zhixing Tan, Shuming Shi, Maosong Sun, and Yang Liu. 2021.
\newblock \href {https://doi.org/10.18653/v1/2021.findings-acl.422} {On the language coverage bias for neural machine translation}.
\newblock In \emph{Findings of the Association for Computational Linguistics: ACL-IJCNLP 2021}, pages 4778--4790.

\bibitem[{Wu et~al.(2021{\natexlab{a}})Wu, Li, Wang, Meng, Qin, Chen, Zhang, Liu et~al.}]{wu2021r}
Lijun Wu, Juntao Li, Yue Wang, Qi~Meng, Tao Qin, Wei Chen, Min Zhang, Tie-Yan Liu, et~al. 2021{\natexlab{a}}.
\newblock R-drop: Regularized dropout for neural networks.
\newblock In \emph{Advances of NeurIPS 2021}, pages 10890--10905.

\bibitem[{Wu et~al.(2021{\natexlab{b}})Wu, Liu, Li, Xu, Chen, Zhang, and Huang}]{Wu:2021:Style}
Xuanxuan Wu, Jian Liu, Xinjie Li, Jinan Xu, Yufeng Chen, Yujie Zhang, and Hui Huang. 2021{\natexlab{b}}.
\newblock \href {https://www.ijcai.org/proceedings/2021/0547.pdf} {Improving stylized neural machine translation with iterative dual knowledge transfer}.
\newblock In \emph{Proceedings of IJCAI 2021}, pages 3971--3977.

\bibitem[{Wu et~al.(2022)Wu, Zhao, Hu, Minervini, Stenetorp, and Riedel}]{Wu:2022:Efficient}
Yuxiang Wu, Yu~Zhao, Baotian Hu, Pasquale Minervini, Pontus Stenetorp, and Sebastian Riedel. 2022.
\newblock An efficient memory-augmented transformer for knowledge-intensive nlp tasks.
\newblock \emph{arXiv preprint arXiv:2210.16773}.

\bibitem[{Yang and Klein(2021)}]{Yang:2021:FUDGE}
Kevin Yang and Dan Klein. 2021.
\newblock \href {https://doi.org/10.18653/v1/2021.naacl-main.276} {{FUDGE}: Controlled text generation with future discriminators}.
\newblock In \emph{Proceedings of NAACL 2021}, pages 3511--3535.

\bibitem[{Yogatama et~al.(2021)Yogatama, de~Masson~d{'}Autume, and Kong}]{Yogatama:2021:Adaptive}
Dani Yogatama, Cyprien de~Masson~d{'}Autume, and Lingpeng Kong. 2021.
\newblock \href {https://doi.org/10.1162/tacl_a_00371} {Adaptive semiparametric language models}.
\newblock \emph{Transactions of the Association for Computational Linguistics}, 9:362--373.

\bibitem[{Zhang et~al.(2018)Zhang, Luan, Sun, Zhai, Xu, Zhang, and Liu}]{Zhang:2018:improving}
Jiacheng Zhang, Huanbo Luan, Maosong Sun, Feifei Zhai, Jingfang Xu, Min Zhang, and Yang Liu. 2018.
\newblock \href {https://doi.org/10.18653/v1/D18-1049} {Improving the transformer translation model with document-level context}.
\newblock In \emph{Proceedings of EMNLP 2018}, pages 533--542.

\bibitem[{Zhang et~al.(2022)Zhang, Guan, Liu, Ding, Lu, Gu, and Gu}]{Zhang:2022:PersonNMT}
Peng Zhang, Zhengqing Guan, Baoxi Liu, Xianghua Ding, Tun Lu, Hansu Gu, and Ning Gu. 2022.
\newblock Building user-oriented personalized machine translator based on user-generated textual content.
\newblock \emph{Proceedings of the ACM on Human-Computer Interaction}, 6(CSCW2):1--26.

\bibitem[{Zheng et~al.(2021)Zheng, Zhang, Huang, Chen, Xie, Luo, and Chen}]{Zheng:2021:Non-parametric}
Xin Zheng, Zhirui Zhang, Shujian Huang, Boxing Chen, Jun Xie, Weihua Luo, and Jiajun Chen. 2021.
\newblock \href {https://doi.org/10.18653/v1/2021.findings-emnlp.358} {Non-parametric unsupervised domain adaptation for neural machine translation}.
\newblock In \emph{Findings of the Association for Computational Linguistics: EMNLP 2021}, pages 4234--4241.

\end{thebibliography}


\section{Appendix}

\subsection{Training Details}
\label{appendix:model-training}

\paragraph{NMT Model Training} We use WMT14~\cite{Bojar:2014:WMT} De-En and WMT20~\cite{Barrault:2020:WMT} Zh-En training data to train NMT models.
For all the involved language pairs (i.e., En-Zh, Zh-En, and De-En), we train the Transformer model using the following hyper-parameters. All the models are optimized by Adam~\cite{kingma2014adam}, with $\beta_1=0.9$, $\beta_2=0.98$ and $\epsilon=10^{-9}$. We train each model for 200K iterations on 4 NVIDIA A100 GPUs, where the training speed is 8.5 iterations per second. We use the learning schedule presented in \citet{Vaswani:2017:Transformer}, with a maximum learning rate of 7e-4 and the warm-up step is 4K. Each mini-batch contains 32K tokens in total. Both the dropout rate and the label smoothing penalty are set to 0.1. During inference, the beam size is 4. For En-Zh and Zh-En, the NMT models have 253.9M parameters. For De-En, the model has 198.3M parameters.
\paragraph{Adapter Training}
We train the proposed memory-augmented adapter for 20K iterations. The maximum learning rate is set to 2e-4 and we restart the learning rate schedule when training adapters. Each mini-batch contains 8K tokens. Each experiment is conducted through a single run. We tune the values of the hyperparameters on the validation set through grid search.

\paragraph{$k$NN Decoding}
To apply $k$NN decoding to our method, we should firstly build a datastore in the same way as that illustrated in \citet{Khandelwal:2021:kNNMT} using our model augmented with the proposed adapters. We use the open-sourced implementation of $k$NN decoding.\footnote{https://github.com/urvashik/knnmt}

\subsection{Details on Baselines}
\label{sec:baseline-details-appendix}

In this subsection, we provide the essential details of the baselines in this work:
\begin{itemize}
    \item {\em Adapter}~\cite{Houlsby:2019:Adapter}: we use the same adapter architecture as presented in \citet{Houlsby:2019:Adapter}. The training hyperparameters are the same as our method, excluding some newly introduced hyperparameters (e.g, $\alpha$ and $\beta$). The default dimension of the hidden layer is set to 64, following \citet{Houlsby:2019:Adapter}. Since our method use more parameters than Adapter, we also train a larger adapter to assimilate the parameter count, whose hidden dimension is set to 512. The bigger adapter still performs worse than our method (16.9 vs. 20.8 in terms of BLEU), indicating that our performance improvement is not totally caused by the larger adapter size.
    \item {\em MT+Rewrite}~\cite{Syed:2020:Adapting}: we train a rewriting model to refine the output of the NMT model. Specifically, we fine-tune a pretrained encoder-decoder model to transfer the model outputs into stylized texts.
    \item {\em $k$NN-MT}~\cite{Khandelwal:2021:kNNMT}: there are mainly three hyperparameters that have a significant impact on performance: $k$, $T$, and $\lambda$. We tune the hyperparameters on the validation set. For style customization, $k$ = 128, $T$ = 30 and $\lambda$ = 0.7 in En-Zh. In Zh-En, $k$ = 16, $T$ = 40 and $\lambda$ = 0.6. For domain customization, $k$ = 16 across all the four domains. $T$ = 4 in IT, Medical, and Law. $T$ = 40 in Koran. $\lambda$ is tuned to be 0.2, 0.3, 0.3, 0.6 in IT, Medical, Law, and Koran, respectively.
    \item {\em DExperts}~\cite{Liu:2021:Dexperts}: we should learn two independent language models, of which one language model serves as an expert and another one is an anti-expert. The expert model is fine-tuned on the user-provided data while the anti-expert is trained on the general domain data. $\alpha$ is tuned on the validation set and the final value is 0.2.
\end{itemize}

\subsection{Case study}

We place some translation examples in Table~\ref{tab:style-case-appendix} to provide a better understanding of the difference between the involved methods. There are 6 sentences from short to long. We can find that our method always outputs better translations. Also, $k$NN-MT is not always better than Adapter, see the 3-rd and 4-th cases.

\citet{Syed:2020:Adapting} believe that the author-style can be understood at three levels, from punctuation, word usage to syntax~\cite{Syed:2020:Adapting}. In our cases, we can find that our method can learn to generate author-style better in different granularities. From case 2, our method correctly translates the phrase ``build the tower'' to ``\begin{CJK}{UTF8}{gbsn}造塔\end{CJK}'' while the other methods translate it to ``\begin{CJK}{UTF8}{gbsn}修建塔\end{CJK}'' or ``\begin{CJK}{UTF8}{gbsn}建造雷峰塔\end{CJK}''. Although the meaning is the same, our translation is closer to the expression style of the original author/user. Similarly, our method properly translates the word ``call'' to ``\begin{CJK}{UTF8}{gbsn}呼唤\end{CJK}'' while the other methods translate it to ``\begin{CJK}{UTF8}{gbsn}打电话\end{CJK}'' or ``\begin{CJK}{UTF8}{gbsn}叫\end{CJK}'' in case 3. Also, our method translates the phrase ``that night'' to ``\begin{CJK}{UTF8}{gbsn}那夜\end{CJK}'' while the other methods translate it to ``\begin{CJK}{UTF8}{gbsn}那一晚\end{CJK}'' or ``\begin{CJK}{UTF8}{gbsn}昨夜\end{CJK}'' in case 4. Furthermore, it can also be easily found that our method can generate similar sentences that have similar syntactic styles. From case 1 and case 6, we can see that the sentence generated by our method is more similar to the reference in terms of sentence segmentation. 

These cases from lexical level to syntactic structure also demonstrate the rationality and effectiveness of our multi-granularity memory design.

\begin{table*}[!t]
  \centering
  \caption{Case study on En-Zh style customization. For clarity, we only choose the representative baselines (i.e., Adapter~\cite{Houlsby:2019:Adapter} and $k$NN-MT~\cite{Khandelwal:2021:kNNMT}) for comparison.}
  \small
  \scalebox{0.92}{
  \begin{tabular}{l m{14.5cm}}
  \toprule
  \bf Source & So biological truth is by no means a talisman for polygamy.  \\
  \cmidrule(lr){1-1} \cmidrule(lr){2-2}
  \bf Adapter & \begin{CJK}{UTF8}{gbsn}
    因此，生物学真理决不是一夫多妻制的护身符。
  \end{CJK} \\
  \cmidrule(lr){1-1} \cmidrule(lr){2-2}
  {\bf $k$NN-MT } & \begin{CJK}{UTF8}{gbsn}
    所以，生物学的真理，决不是一夫多妻制的护身的挡牌。
  \end{CJK} \\
  \cmidrule(lr){1-1} \cmidrule(lr){2-2}
  \bf Ours & \begin{CJK}{UTF8}{gbsn}
    所以生物学上的道理，决不是一夫多妻的护身符。
  \end{CJK} \\
  \cmidrule(lr){1-1} \cmidrule(lr){2-2}
  {\bf Reference} & \begin{CJK}{UTF8}{gbsn}
    所以生物学的真理，决非多妻主义的护符。
  \end{CJK} \\
  \midrule
  \midrule
  \bf Source & Could it be that when he built the tower, he didn't think that the tower would fall down after all. \\
  \cmidrule(lr){1-1} \cmidrule(lr){2-2}
  \bf Adapter & \begin{CJK}{UTF8}{gbsn}
    难道他修建塔的时候，总不认为塔会倒塌吗。
  \end{CJK}\\
  \cmidrule(lr){1-1} \cmidrule(lr){2-2}
  \bf $k$NN-MT & \begin{CJK}{UTF8}{gbsn}
    倘若他建造雷峰塔的时候，他没有想到那塔终究会倒塌。
  \end{CJK} \\
  \cmidrule(lr){1-1} \cmidrule(lr){2-2}
  \bf Ours & \begin{CJK}{UTF8}{gbsn}
    难道他造塔的时候，总不觉得那塔到底要倒下去么。
  \end{CJK} \\
  \cmidrule(lr){1-1} \cmidrule(lr){2-2}
  \bf Reference & \begin{CJK}{UTF8}{gbsn}
    莫非他造塔的时候，竟没有想到塔是终究要倒的么。
  \end{CJK} \\
  \midrule
  \midrule
  \bf Source & It was the morning of the fifth day that everyone dragged him up early in the morning and stood on the shore listening to the call. \\
  \cmidrule(lr){1-1} \cmidrule(lr){2-2}
  \bf Adapter & \begin{CJK}{UTF8}{gbsn}
    这是第五天早晨，大家一大早就把他拖起来，站在岸上听着叫。
  \end{CJK}\\
  \cmidrule(lr){1-1} \cmidrule(lr){2-2}
  \bf $k$NN-MT & \begin{CJK}{UTF8}{gbsn}
    第二天的早晨，大家一大早把他拖起来，站在岸上听听差打电话。
  \end{CJK} \\
  \cmidrule(lr){1-1} \cmidrule(lr){2-2}
  \bf Ours & \begin{CJK}{UTF8}{gbsn}
    这是第五天的早晨，大家早把他拖起来，站在岸上听着呼唤。
  \end{CJK} \\
  \cmidrule(lr){1-1} \cmidrule(lr){2-2}
  \bf Reference & \begin{CJK}{UTF8}{gbsn}
    就是这第五天的早晨，大家一早就把他拖起来，站在岸上听呼唤。
  \end{CJK} \\
  \midrule
  \midrule
  \bf Source & Really, until now, I really haven't eaten the good beans like that night, and I haven't seen the good show like that night anymore. \\
  \cmidrule(lr){1-1} \cmidrule(lr){2-2}
  \bf Adapter & \begin{CJK}{UTF8}{gbsn}
    真的，到目前为止，我实在没有吃过那一晚这样的好豆子，我再也没有见过那晚这样的好秀了。
  \end{CJK}\\
  \cmidrule(lr){1-1} \cmidrule(lr){2-2}
  \bf $k$NN-MT & \begin{CJK}{UTF8}{gbsn}
    真的，直到现在，我实在没有吃过昨夜的豆子，我也从来没见过这样的好东西。
  \end{CJK} \\
  \cmidrule(lr){1-1} \cmidrule(lr){2-2}
  \bf Ours & \begin{CJK}{UTF8}{gbsn}
    真的，直到现在，我实在没有吃过那夜那样好的豆子了，我也再没有见过那夜那样的好节目了。
  \end{CJK} \\
  \cmidrule(lr){1-1} \cmidrule(lr){2-2}
  \bf Reference & \begin{CJK}{UTF8}{gbsn}
    真的，一直到现在，我实在再没有吃到那夜似的好豆，也不再看到那夜似的好戏了。
  \end{CJK} \\
  \midrule
  \midrule
  \bf Source & This affair happened at the turn of winter and spring. The wind was not so cold anymore, and I wandered outside for a longer time; by the time I got home, it had been probably already dark. \\
  \cmidrule(lr){1-1} \cmidrule(lr){2-2}
  \bf Adapter & \begin{CJK}{UTF8}{gbsn}
    这事发生在冬春交替的时候，风不再那么冷了，我在外面漫游了更长的时间；到我到家的时候，大概是已经暗了。
  \end{CJK}\\
  \cmidrule(lr){1-1} \cmidrule(lr){2-2}
  \bf $k$NN-MT & \begin{CJK}{UTF8}{gbsn}
    这事发生在冬或今年春末，微风不再那么冷，我徘徊了大半天，到我回家的时候，天气大概已经很深了。
  \end{CJK} \\
  \cmidrule(lr){1-1} \cmidrule(lr){2-2}
  \bf Ours & \begin{CJK}{UTF8}{gbsn}
    这事发生于冬春之交，风已不再那么冷，我在外面徘徊了更长的时间；到我到家的时候，大概已经天黑了。
  \end{CJK} \\
  \cmidrule(lr){1-1} \cmidrule(lr){2-2}
  \bf Reference & \begin{CJK}{UTF8}{gbsn}
    这是冬春之交的事，风已没有这么冷，我也更久地在外面徘徊；待到回家，大概已经昏黑。
  \end{CJK} \\
  \midrule
  \midrule
  \bf Source & For example, to build a railway, if we tell them how beneficial this thing is, they will never listen. If we, according to a myth, tell that previously a great immortal pushed a wheelbarrow over the rainbow, and now we imitate him to build a road, then everything can be done. \\
  \cmidrule(lr){1-1} \cmidrule(lr){2-2}
  \bf Adapter & \begin{CJK}{UTF8}{gbsn}
    举例来说，要修建铁路，如果我们告诉他们这事是何等的益处，他们就决不肯听从，如果我们，根据一个神话，告诉以前一个伟大的不朽的推车推着彩虹之上，现在我们模仿他修筑一条路，那么一切都可以完成了。
  \end{CJK}\\
  \cmidrule(lr){1-1} \cmidrule(lr){2-2}
  \bf $k$NN-MT & \begin{CJK}{UTF8}{gbsn}
    譬如说，如果我们告诉他们，铁路的建设是多么有益，他们决不听，如果我们根据神话说，先前一个大不朽的独轮车从彩虹上推过，现在我们模仿他造路，便可以做点事了。
  \end{CJK} \\
  \cmidrule(lr){1-1} \cmidrule(lr){2-2}
  \bf Ours & \begin{CJK}{UTF8}{gbsn}
    譬如造铁路罢，倘告诉他们这东西有多大益处，他们便决不肯听话，倘使我们据传说，说先前一个伟大的不朽的车手推着彩虹，现在就模仿他来造一条路，那么，一切便都可以做。
  \end{CJK} \\
  \cmidrule(lr){1-1} \cmidrule(lr){2-2}
  \bf Reference & \begin{CJK}{UTF8}{gbsn}
    譬如要造一条铁路，倘若对他们说这事如何有益，他们决不肯听；我们如果根据神话，说从前某某大仙，曾推着独轮车在虹霓上走，现在要仿他造一条路，那便无所不可了。
  \end{CJK} \\
  \bottomrule
  \end{tabular}}
  \label{tab:style-case-appendix}
\end{table*}

\subsection{Application to Larger Model}

We also conduct experiments on a model of a larger scale, whose hidden size is 1024 and parameter size is 596.0M. On the test set of En-Zh style customization, our memory-augmented adapter (without $k$NN decoding) outperforms Adapter~\cite{Houlsby:2019:Adapter} by 2.9 BLEU and $k$NN-MT~\cite{Khandelwal:2021:kNNMT} by 2.2 BLEU. This demonstrates that our method is also effective when the model size is larger. How to apply our method to larger models deserves further exploration.

\end{document}